\documentclass[10pt,twocolumn,letterpaper]{article}

\usepackage{iccv}
\usepackage{times}
\usepackage{epsfig}
\usepackage{graphicx}
\usepackage{amsmath}
\usepackage{amssymb}

\usepackage{subfigure}

\usepackage{interval}

\usepackage{caption}

\usepackage{pifont}

\usepackage{makecell}
\usepackage{booktabs}
\usepackage{multirow}
\usepackage{siunitx}

\usepackage[T1]{fontenc}
\usepackage{fix-cm}

\newcolumntype{?}{!{\vrule width 1pt}}
\newcolumntype{C}[1]{>{\centering}m{#1}}

\newcolumntype{X}{@{\hskip\tabcolsep\vrule width 1.5pt\hskip\tabcolsep}}

\newcommand{\myfiguresixcol}[1]{
\begin{minipage}[b]{.14\textwidth}
\includegraphics[width=1.1\linewidth]{#1}
\end{minipage}
}

\usepackage[pagebackref=true,breaklinks=true,letterpaper=true,colorlinks,bookmarks=false]{hyperref}

 \iccvfinalcopy 

\def\iccvPaperID{883} 

\ificcvfinal\fi
\begin{document}

\title{Am I a Baller? Basketball Performance Assessment \\ from First-Person Videos}

\author{Gedas Bertasius$^{1}$, Hyun Soo Park$^2$, Stella X. Yu$^3$, Jianbo Shi$^1$\\
$^1$University of Pennsylvania, $^2$University of Minnesota, $^3$UC Berkeley ICSI\\
{\tt\small \{gberta,jshi\}@seas.upenn.edu} \ \ \ {\tt\small hspark@umn.edu}  \ \ \ {\tt\small stella.yu@berkeley.edu}
}


\maketitle

\begin{abstract}

This paper presents a method to assess a basketball player's performance from his/her first-person video. A key challenge lies in the fact that the evaluation metric is highly subjective and specific to a particular evaluator. We leverage the first-person camera to address this challenge. The spatiotemporal visual semantics provided by a first-person view allows us to reason about the camera wearer's actions while he/she is participating in an unscripted basketball game. Our method takes a player's first-person video and provides a player's performance measure that is specific to an evaluator's preference.

To achieve this goal, we first use a convolutional LSTM network to detect atomic basketball events from first-person videos. Our network's ability to zoom-in to the salient regions addresses the issue of a severe camera wearer's head movement in first-person videos. The detected atomic events are then passed through the Gaussian mixtures to construct a highly non-linear visual spatiotemporal basketball assessment feature. Finally, we use this feature to learn a basketball assessment model from pairs of labeled first-person basketball videos, for which a basketball expert indicates, which of the two players is better. 

We demonstrate that despite not knowing the basketball evaluator's criterion, our model learns to accurately assess the players in real-world games. Furthermore, our model can also discover basketball events that contribute positively and negatively to a player's performance.

\end{abstract}

\vspace{-0.5cm}

\section{Introduction}

A gifted offensive college basketball player, Kris Jenkins (Villanova), made a three point buzzer beater against UNC (2015-2016 season), and recorded one of the greatest endings in NCAA championship history. He was arguably one of the best players in the entire NCAA tournament. A question is ``what makes him stand out from his peer players?''. His stats, e.g., average points and rebounds per game, can be a measure to evaluate his excellence. However, these measures do not capture every basketball aspect that a coach may want to use for assessing his potential impact in the future team, which is difficult to measure quantitatively. NBA coaches and scouts are eager to catch every nuance of a basketball player's abilities by watching a large number of his basketball videos.

\captionsetup{labelformat=default}
\captionsetup[figure]{skip=10pt}

\begin{figure}
\begin{center}
   \includegraphics[width=1.025\linewidth]{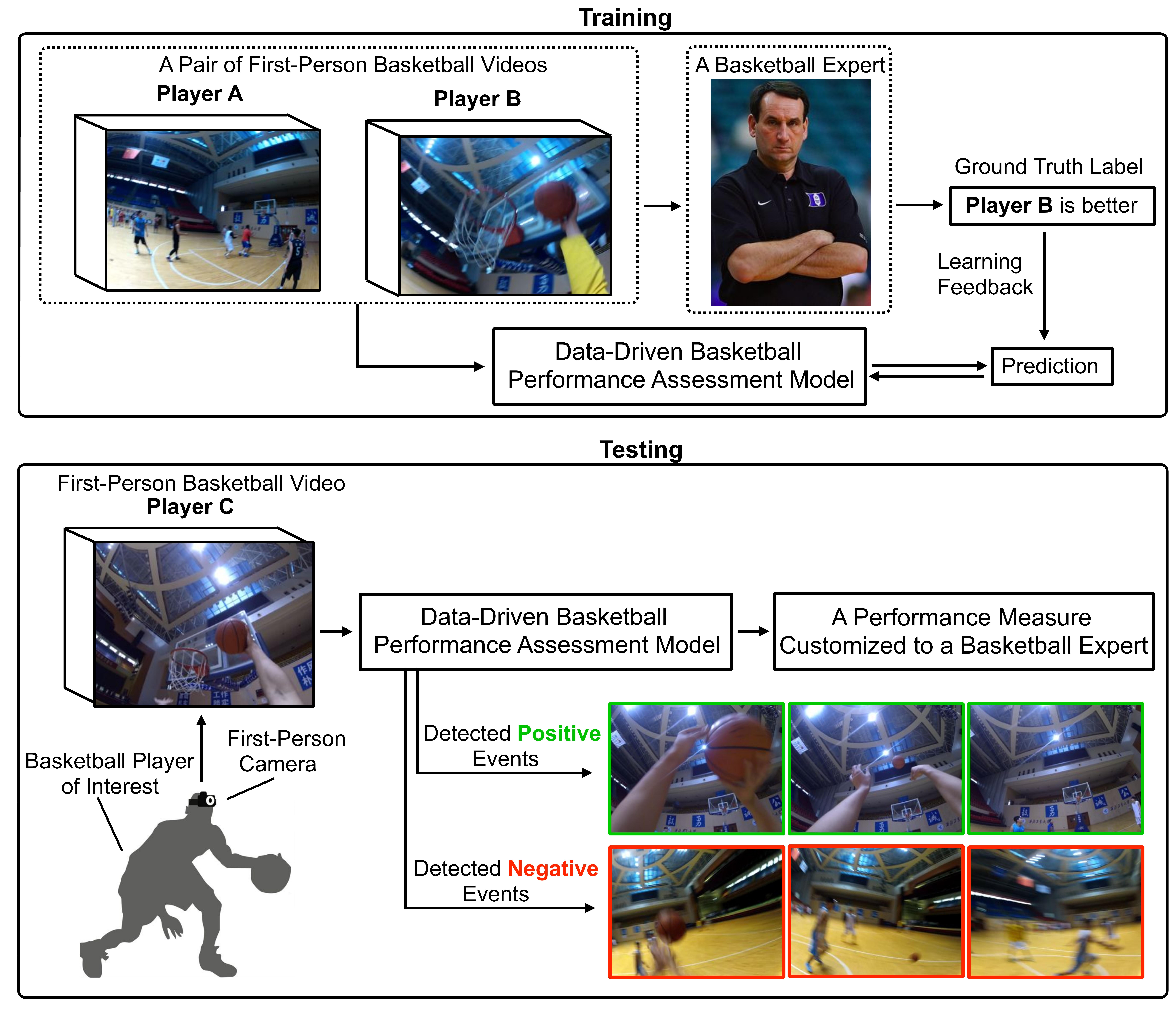}
\end{center}
\vspace{-0.4cm}
   \caption{Our goal is to assess a basketball player's performance based on an evaluator's criterion from an unscripted  first-person basketball video of a player. During training, we learn such a model from the pairs of weakly labeled first-person basketball videos. During testing, our model predicts a performance measure customized to a particular evaluator from a first-person basketball video. Additionally, our model can also discover basketball events that contribute positively and negatively to a player's performance.\vspace{-0.4cm}}
   
\label{main_fig}
\end{figure}

Now consider a college recruitment process where there is a massive number of high school players. In such conditions, the searching task for the best players becomes much more challenging, more expensive and also more labor intense. More importantly, the recruiters need to measure and evaluate a sequence of atomic decision makings, e.g., when does a player shoot, whether he makes a shot, how good is his passing ability, etc. There exists neither universal measure nor golden standard to do this, i.e., most scouts and coaches have their own subjective evaluation criterion.

In this paper, we address a problem of computational basketball player assessment customized to a coach's or scout's evaluation criterion. Our conjecture is that a first-person video captures a player's basketball actions and his/her basketball decision making in a form of the camera motion and visual semantics of the scene. A key challenge of first-person videos is that it immediately violates primary assumptions made for third-person recognition systems: first-person videos are highly unstable and jittery and visual semantics does not appear as iconic as in third-person~\cite{imagenet_cvpr09}.

Our first-person approach innovates the traditional assessment methods, e.g., watching hours of third-person videos taken by non professional videographers and assessing the players in them. In contrast, a first-person video records what the player sees, which directly tells us what is happening to the player himself, e.g., the body pose of a point guard who is about to pass at HD resolution while a third-person video produces a limited visual access to such subtle signals. Furthermore, the 3D camera egomotion of the first person video reflects the decision making of how the player responds to the team configuration, e.g., can I drive towards the basket and successfully finish a layup? Finally, a first-person camera eliminates the tracking and player association tasks of the third-person video analysis, which prevents applications of computational approaches for amateur games\footnote{Usage of commercial tracking systems using multiple calibrated cameras is limited due to a high cost~\cite{stats_vu}}.

Our system takes a first-person video of basketball players and outputs a basketball assessment metric that is specific to an evaluator's preference. The evaluator provides the comparative weak labels of the performance of the players, e.g., the player $\mathsf{A}$ is better than $\mathsf{B}$ based on his own subjective criteria. 

Our method first uses a convolutional LSTM to detect atomic basketball events from a first-person video. Our network's ability to localize the most informative regions in a first-person image, is essential for first-person videos where the camera undergoes severe head movement, which causes videos to be blurry. These atomic events are then passed through the Gaussian mixtures to produce a highly non-linear visual spatiotemporal basketball assessment feature. Finally, our basketball assessment model is learned from the pairs of labeled first-person basketball videos by minimizing a hinge loss function. We learn such a basketball skill assessment model from our new $10.3$ hour long first-person basketball dataset that captures $48$ distinct college level basketball players in an unscripted basketball game.

\textbf{Impact} Ample money and effort have been invested in recruiting, assessing, and drafting basketball players every year. However, limited progress has been made on developing computational models that can be used to automatically assess an athlete's performance in a particular sport~\cite{10.1109/ICDM.2014.106,open_shot}. As wearable technology advances, cameras can be non-invasively worn by players, which delivers a vivid sense of their dynamics, e.g., Spanish Liga ACB has demonstrated a possibility of a jersey camera that allows you to put yourself in the court~\cite{acb}. This trend will open up a new opportunity to share experiences and evaluate performance across players in different continents without bias and discrimination. Our work takes a first step towards enabling a computational analysis for such first-person data.   

\textbf{Contribution} To the best of our knowledge, this is the first paper that addresses practical behavioral assessment tasks using first-person vision specific to an evaluator's preference. The core technical contributions of the paper include 1) a basketball assessment model that assesses the players based an an evaluator's assessment criterion, which we learn from the pairs of weakly labeled first-person basketball videos; 2) a predictive network that learns the visual semantics of important actions and localizes salient regions of first-person images to handle unstable first-person videos and 3) a new $10.3$ hour long first-person basketball video dataset capturing $48$ players in an unscripted basketball game.

\section{Related Work}

\noindent\textit{Talent wins games, but teamwork and intelligence wins championships.}  --- Michael Jordan\\

Accurate diagnosis and evaluation of athletes is a key factor to build a synergic teamwork. However, it is highly subjective and task dependent, and the psychological and financial cost of such process is enormous. A large body of sport analytics and kinesiology has studied a computational approaches to provide a quantitative measure of the performance~\cite{10.1109/ICDM.2014.106,open_shot,tennis,ghosting,nba_strat}. 

Kinematic abstraction (position, orientation, velocity, and trajectory) of the players offers a global centric representation of team behaviors, which allows a detailed analysis of the game such as the probability of shoot success, rebound, and future movement prediction~\cite{tennis,ghosting,park_cvpr:2017}. Not only an individual performance, but also team performance can be measured through the kinematic abstraction~\cite{nba_strat,open_shot}. 

These kinematic data are often obtained by multiple third-person videos~\cite{stats_vu, ghosting, open_shot} where the players and ball are detected using recognition algorithms combined with multiple view geometry~\cite{Hartley2004}. Tracking and data association is a key issue where the role of the players provides a strong cue to disambiguate appearance based tracking~\cite{10.1109/CVPR.2013.349}. Events such as ball movement, can be also recognized using a spatiotemporal analysis~\cite{Maksai_2016_CVPR}. As players behave strategically and collectively, their group movement can be predicted~\cite{Kim:2012:GP-ROI} and the ball can be localized without detection. Various computational models have been used for such tasks, e.g., Dynamic Bayesian Network~\cite{Swears+Hoogs+Ji+Boyer2014}, hierarchical LSTM~\cite{msibrahi16deepactivity}, attention based LSTM~\cite{ramanathan_cvpr16} learned from a large collection of third-person videos.   

Unlike third-person videos, first-person cameras closely capture what the players see. Such property is beneficial to understand activities highly correlated with visual attention, e.g., object manipulation and social communications. Important objects to the camera wearer are detected and segmented~\cite{DBLP:journals/ijcv/LeeG15,BMVC.28.30,conf/cvpr/RenG10,conf/cvpr/FathiRR11,DBLP:journals/corr/BertasiusPYS16}, which can be used to compress life-log videos~\cite{DBLP:journals/ijcv/LeeG15,Lu:2013:SSE:2514950.2516026}. As visual attention is also related with the intent of the camera wearer, her/his future movement can be predicted~\cite{park_ego_future}. Beyond individual behaviors, joint attention is a primary indicator of social interactions, which can be directly computed from first-person videos~\cite{Fathi_socialinteractions:,park_nips:2012}, and further used for human-robot interactions~\cite{Ryoo:2015:RAP:2696454.2696462,DBLP:journals/corr/GoriAR15}.   

In sports, the complex interactions with a scene in first-person videos can be learned through spatiotemporal visual patterns. For instance, the scene can tell us about the activity~\cite{conf/cvpr/KitaniOSS11} and the egomotion can tell us about the physical dynamics of activity~\cite{park_force}. Joint attention still exists in team sports which can be described by the team formation~\cite{park_cvpr:2015} and future behaviors~\cite{park_cvpr:2017}. 

Unlike previous work that mainly focuses on recognizing and tracking objects, activities, and joint attention, we take one step further: performance assessment based on the evaluator's preference. We introduce a computational model that exhibits strong predictive power when applied on the real world first-person basketball video data.

   \begin{figure*}
\begin{center}
   \includegraphics[width=1\linewidth]{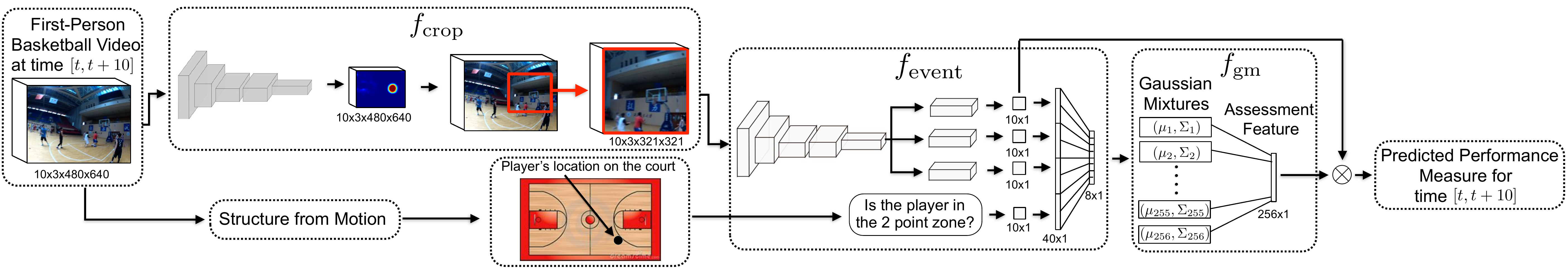}
\end{center}
\vspace{-0.6cm}
\caption{A detailed illustration of our basketball assessment prediction scheme. Given a video segment from time interval $[t,t+10]$, we first feed it through a function $f_{\rm crop}$, which zooms-in to the relevant parts of a video. We then apply  $f_{\rm event}$ to predict $4$ atomic basketball events from a zoomed-in video and a player's $(x,y)$ location on the court. We then feed these predictions through a Gaussian mixture function  $f_{\rm  gm}$, which produces a highly non-linear visual spatiotemporal assessment feature. Finally, we use this feature to compute a player's assessment measure by multiplying it with linear weights learned from the data, and with a predicted relevance indicator for a given video segment.\vspace{-0.4cm}}



\label{test_arch_fig}
\end{figure*}

\section{Basketball Performance Assessment Model} 
\label{model_sec}

We define a measure of performance assessment using a first-person video:
\begin{align}
S(\mathcal{V}) = \frac{\sum_{t=1}^T p^{(1)}_t \mathbf{w}^\mathsf{T}\boldsymbol{\phi} (\mathbf{V}_t, \mathbf{x})}{\sum_{t=1}^T p^{(1)}_t} \label{Eq:assessment}
\end{align}
where $\mathcal{V}$ is a first-person video of $T$ number of frames, $\phi$ is a visual spatiotemporal basketball assessment feature, and $\mathbf{w}$ is a weight vector of performance regressor. $\mathbf{V}_t \subset \mathcal{V}$ is a segmented video starting at the $t^{\rm th}$ frame with a fixed length, $T_s$. $p^{(1)}_t \in [0, 1]$ is a relevance of $\mathbf{V}_t$ to evaluate a given player's performance. $\mathbf{x} \in \mathbb{R}^2$ is the 2D coordinate of the basketball player, i.e., the projection of 3D camera pose computed by structure from motion~\cite{Hartley2004} onto the canonical basketball court.  In Figure~\ref{test_arch_fig}, we provide a detailed illustration of our basketball assessment prediction framework.

\subsection{Visual Spatiotemporal Assessment Feature}

Our first goal is to use a first-person basketball video to build a powerful feature representation that could be used for an effective player's performance assessment. We identify three key challenges related to building such a representation from first-person basketball videos: 1) our system needs to handle severe camera wearer's head motion, 2) we need to have an interpretable basketball representation in terms of its atomic events, and 3) our feature representation has to be highly discriminative for a player's performance prediction task. 

To address these problems, we propose to represent the visual feature of the segmented video, $\mathbf{V}_t$, as follows, where each function below addresses one of the listed challenges:
\begin{align}
    \boldsymbol{\phi}(\mathbf{V}_t,\mathbf{x}) = f_{\rm gm} \left( f_{\rm event} \left( f_{\rm crop} \left(\mathbf{V}_t\right), \mathbf{x}\right)\right),
\end{align}
where $f_{\rm crop}$ is a function that handles a severe camera wearer's head motion by producing a cropped video by zooming in on the important regions, $f_{\rm event}$ is a function that computes the probability of atomic basketball events, and $f_{\rm gm}$ is a Gaussian mixture function that computes a highly non-linear visual feature of the video.

\textbf{Zooming-In.} A key property of $f_{\rm crop}$ is the ability to zoom-in to relevant pixels which allows to learn an effective visual representation for the basketball performance assessment. Using this regional cropping, we minimize the effect of jittery and unstable nature of first person videos that causes larger variation of visual data. In our experimental section, we demonstrate that using $f_{\rm crop}$ in our model substantially improves the prediction performance. Thus, initially we process a first-person video to produce a cropped video:
\begin{align}
    \overline{\mathbf{V}}_t = f_{\rm crop} (\mathbf{V}_t; \mathbf{w}_{\rm crop}),\nonumber
\end{align}
where $f_{\rm crop}$ is parametrized by $\mathbf{w}_{\rm crop}$, $\overline{\mathbf{V}}_t$ is the cropped video with fixed size $C_w\times C_w\times 3 \times T_s$, and $C_w$ is the width and height of the cropping window.    


We predict the center of the cropping window by learning $\mathbf{w}_{\rm crop}$ using a fully convolutional network~\cite{DBLP:journals/corr/ChenYWXY15}. To do this, we train the network to predict the location of a ball, which is typically where most players are looking at. Afterwards, for each frame in a video, we compute a weighted average of $XY$ location coordinates weighted by the detected ball probabilities and then crop a fixed size patch around such a weighted average location. We illustrate some of the qualitative zoom-in examples in Figure~\ref{best_worst_fig}.


\textbf{Atomic Basketball Event Detection.}   To build an interpretable representation in terms of atomic basketball events, we predict basketball events of 1) sombeody shooting a ball,  2) the camera wearer possessing the ball, and 3) a made shot respectively. Note that the cropped video focuses on the ball and its visual context, which allows to learn the visual semantics of each atomic event more effectively. To do this we use a multi-path convolutional LSTM network, where each pathway predicts its respective atomic basketball event. We note that such a multi-path architecture is beneficial as it allows each pathway to focus on learning a single atomic basketball concept. In contrast, we observed that training a similar network with a single pathway failed to produce accurate predictions for all three atomic events. Given a cropped video, our multi-path network is jointly trained to minimize the following cross-entropy loss:

\begin{equation*}
\begin{split}
    \mathcal{L}_{\rm event} = -\sum_{t=1}^{T_s} \sum_{b=1}^{3} y^{(b)}_t \log p^{(b)}_{t}+(1-y^{(b)}_t) \log \left(1-p^{(b)}_{t}\right), \nonumber
\end{split}
\end{equation*}

where $p^{(b)}_t$ depicts a network's prediction for an atomic basketball event $b$ at a time step $t$; $y^{(b)}_t \in \{0,1\}$ is a binary atomic basketball event ground truth value for frame $t$ and basketball event $b$. 

We also note that because many important basketball events occur when somebody shoots the ball~\cite{nba_savant,sloan}, the detected probability $p^{(1)}_t$ is also later used in Equation~(\ref{Eq:assessment}), as a relevance indicator for each video segment, $\mathbf{V}_t$. 

As our fourth atomic basketball event $p^{(4)}_t$, we use a binary value indicating whether a player is in the 2 point or 3 point zone, which is obtained from a player's $(x,y)$ location coordinates on the court. 


We then split each of the $4$ basketball event predictions in half across the temporal dimension, and perform temporal max pooling for each of the $8$ blocks. All the pooled values are then concatenated into a single vector $\mathbf{b}_t$:

\vspace{-0.3cm}
\begin{align}
    \mathbf{b}_t =  f_{\rm event} (\overline{\mathbf{V}}_t,\mathbf{x};\mathbf{w}_{\rm event}) \nonumber
\end{align}

\textbf{Gaussian Mixtures.} To build a representation that is discriminative, and yet generalizable, we construct a highly non-linear feature that works well with a linear classifier. To achieve these goals we employ Gaussian mixtures, that transform the atomic basketball event feature, into a complex basketball assessment feature, which we will show to be very effective in our assessment model. Formally, given a vector $\mathbf{b}_t$ over $T_s$, we compute the visual spatiotemporal assessment features for a given video segment as:
\begin{align}
\boldsymbol{\phi}_t = f_{\rm gm} \left(\mathbf{b}_t;\{\boldsymbol{\mu}_n,\boldsymbol{\Sigma}_n\}_{n=1}^N\right)\nonumber
\end{align}

   \begin{figure*}
\begin{center}
   \includegraphics[width=1\linewidth]{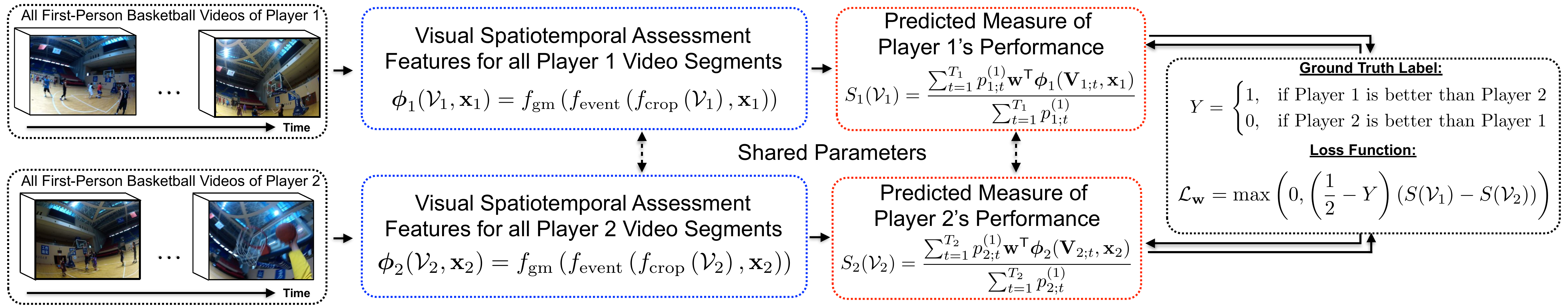}
\end{center}
\vspace{-0.6cm}
\caption{An illustration of of our training procedure to learn the linear weights $w$ that are used to assess a given basketball player's performance. As an input we take a pair of labeled first-person basketball videos with a label provided by a basketball expert indicating, which of the two players is better. Then, we compute visual spatiotemporal basketball assessment features for all input video segments, and use them to learn weights $w$ by minimizing our formulated hinge loss function. \vspace{-0.4cm}} 

\label{train_arch_fig}
\end{figure*}

where $f_{\rm gm}$ is parametrized by Gaussian mixtures, $\{\boldsymbol{\mu}_n,\boldsymbol{\Sigma}_n\}_{n=1}^N$, and $N$ is the number of mixtures. Each mixture $j$ is defined by a function $z(y^{(1)}_{t_1},y^{(1)}_{t_2}, \hdots , y^{(4)}_{t_1},y^{(4)}_{t_2})=j$. Here $y^{(i)}_{t_1}, y^{(i)}_{t_2} \in \{0,1\}$ refer to the binary ground truth values associated with an atomic basketball event $i \in \{1,2,3,4\}$; the index $t_1$ indicates the first half of an input video segment, whereas $t_2$ indicates the second half.  Every possible combination of these values define one of the $2^8=256$ Gaussian mixtures. We learn the parameters of each Gaussian mixture using maximum likelihood from the training data with diagonal covariances.


\subsection{Basketball Assessment Prediction}

We learn a linear weight $\mathbf{w}$ in Equation~(\ref{Eq:assessment}) based on the comparative assessment of players provided by a former professional basketball player in Section~\ref{data_sec}. We minimize the following hinge loss:
\begin{align}
\mathcal{L}_\mathbf{w} = \sum_{i=1}^D \max \left(0, \left(\frac{1}{2}-Y_i\right)\left(S(\mathcal{V}_1^i)-S(\mathcal{V}_2^i)\right)\right),
\end{align}


where $Y_i=1$ if a basketball expert declared Player 1 to be better than Player 2; otherwise $Y_i=0$ . $S(\mathcal{V}_1^i),S(\mathcal{V}_2^i)$ depict our predicted performance measure for Players 1, and 2 respectively, $\mathcal{V}_1^i$ and $\mathcal{V}_2^i$ are the first-person basketball videos of Player 1 and Player 2 respectively, and $D$ is the number of data points. Then based on Equation~\ref{Eq:assessment}, we can compute the subgradients of this loss function with respect to $w$ and find $w$ by minimizing it via a standard gradient descent. In Figure~\ref{train_arch_fig}, we provide an illustration of such a learning framework.

\textbf{Why Linear Classifier?} We only have $250$ labeled pairs for learning the weights, which is a small amount of training data. Thus, making a classifier more complex typically results in a severe overfitting. Through our experiments, we discovered that linear weights work the best.

\subsection{Implementation Details}

For all of our experiments involving CNNs, we used a Caffe library~\cite{jia2014caffe}. Both networks were based on DeepLab's~\cite{DBLP:journals/corr/ChenYWXY15} architecture and were trained for $4000$ iterations with a learning rate of $10^{-8}$, $0.9$ momentum, the weight decay of $5 \cdot 10^{-5}$, and $30$ samples per batch. The LSTM layers inside the atomic basketball event network spanned $10$ consecutive frames in the video input. Each pathway in the atomic basketball event network was composed of two $1024$ dimensional convolution layers with kernel size $1 \times 1$ and a $1024$ dimensional LSTM layer. The networks were trained using standard data augmentation. To learn the weights $w$ we used a learning rate of $0.001$ and ran gradient descent optimization for $100$ iterations.

\section{First-Person Basketball Dataset}
\label{data_sec}

We present a first person basketball dataset composed of 10.3 hours of videos with 48 college players. Each video is about 13 minutes long captured by GoPro Hero 3 Black Edition mounted with a head strip. It is recorded at 1280$\times$960 with 100 fps. We record $48$ videos during the two days, with a different group of people playing each day. We use $24$ videos from the first day for training and $24$ videos from the second day for testing. We extract the video frames at $5$ fps to get $98,452$ frames for training, and $87,393$ frames for testing.

We ask a former professional basketball player (played in an European national team) to label which player performs better given a pair of first-person videos. Total 500 pairs are used: 250 for training and 250 for testing. Note that there were no players overlapping between the training and testing splits. 

We also label three simple basketball events: 1) somebody shooting a ball, 2) the camera wearer possessing the ball, and 3) a made shot. These are the key atomic events that drive a basketball game. In total, we obtain  $3,734$, $4,502$, and $2,175$ annotations for each of these three events respectively.

Furthermore, to train a ball detector we label the location of a ball at $5,073$ images by clicking once on the location. We then place a fixed sized Gaussian around those locations and use it as a ground truth label.


 \setlength{\tabcolsep}{3.5pt}

   \begin{table}[t]
   \footnotesize
    \begin{center}
    \begin{tabular}{ c  | c |  c |  c |  c |}
    \cline{2-5}
    & \multicolumn{4}{ c |}{Atomic Events} \\
    \cline{2-5}
    &  \multicolumn{1}{ c |}{$p^{(1)}$} &  \multicolumn{1}{ c |}{$p^{(2)}$}  &  \multicolumn{1}{ c |}{$p^{(3)}$} &  \multicolumn{1}{ c |}{mean}\\ \cline{1-5}
        \multicolumn{1}{| c |}{Tran et al.~\cite{Tran:2015:LSF:2919332.2919929}}  & 0.312 & 0.428 & 0.193 & 0.311 \\
       \multicolumn{1}{| c |}{Singh et al~\cite{Singh_2016_CVPR}} & 0.469 & 0.649 & 0.185 & 0.434 \\
       \multicolumn{1}{| c |}{Bertasius et al~\cite{gberta_2017_RSS}} & 0.548 & 0.723 & 0.289 & 0.520 \\
       \multicolumn{1}{| c |}{Ma et al~\cite{ma2016going}} & 0.622 & 0.718 & 0.364 & 0.568 \\ \hline
       \multicolumn{1}{| c |}{Ours: no LSTM \& no zoom-in} & 0.711 & 0.705 & 0.192 & 0.536 \\ 
       \multicolumn{1}{| c |}{Ours: no zoom-in} & 0.693 & 0.710 & 0.248 & 0.550 \\ 
       \multicolumn{1}{| c |}{Ours: single path} & 0.678 & 0.754 & 0.308 & 0.580 \\ 
       \multicolumn{1}{| c |}{Ours: no LSTM} & 0.718 & 0.746 &\bf  0.397 & 0.620 \\ \hline
       \multicolumn{1}{| c |}{Ours} & \bf 0.724 & \bf 0.756 & 0.395 & \bf 0.625 \\ \hline
    \end{tabular}
    \end{center}
    \vspace{-0.2cm}
    \caption{The quantitative results for atomic basketball event detection on our first-person basketball dataset according to max F-score (MF) metric. These results show that our  method 1) outperforms prior first-person methods and 2) that each component plays a critical role in our system.\vspace{-0.4cm}}
    \label{att_table}
   \end{table}

   \setlength{\tabcolsep}{1.5pt}

%

\begin{figure*}[t]
\centering

\subfigure[A Pair of Players  \#1]{\label{good_spatial_fig}\includegraphics[height=0.15\textheight]{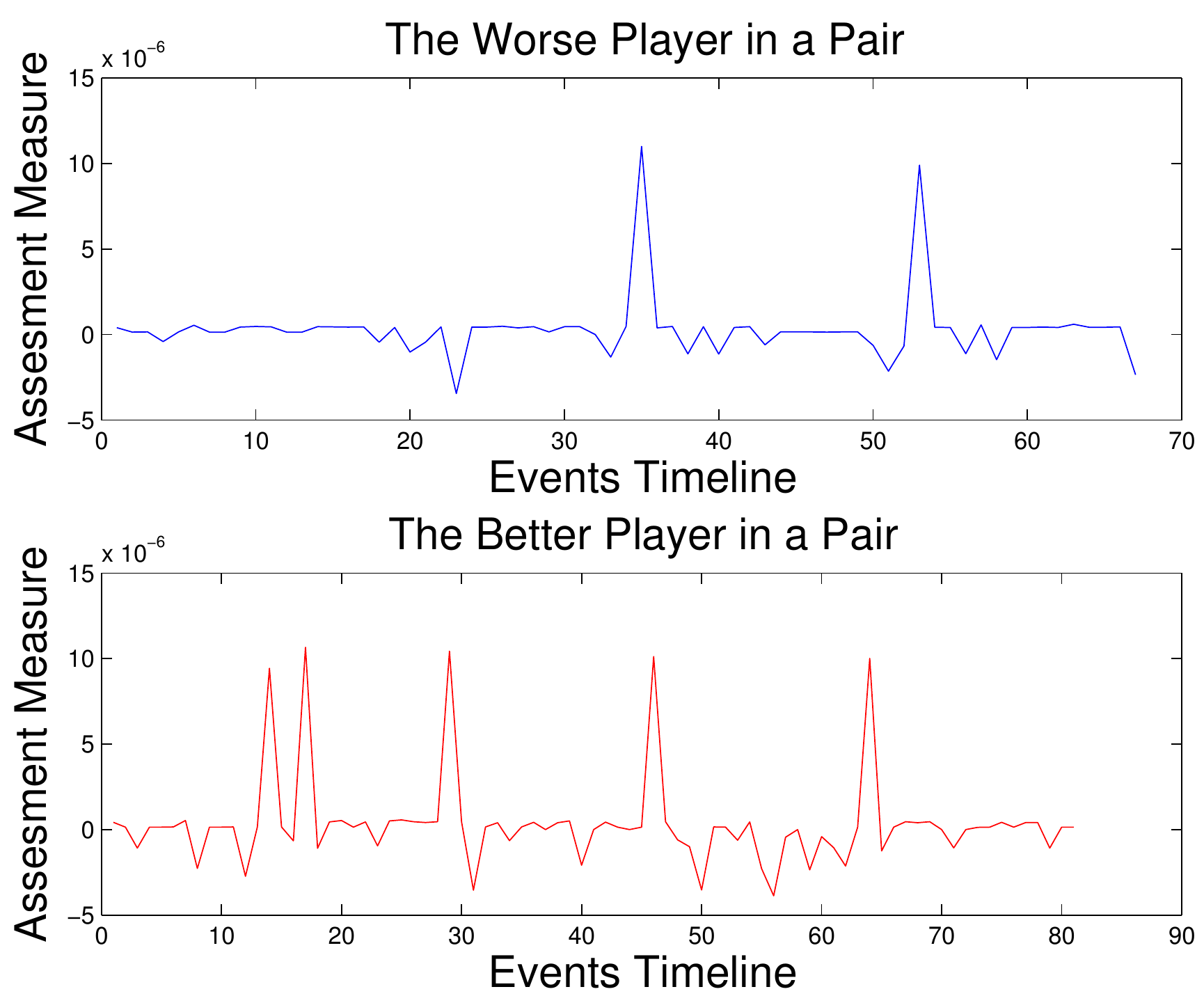}}~
\subfigure[A Pair of Players  \#2]{\label{bad_spatial_fig}\includegraphics[height=0.15\textheight]{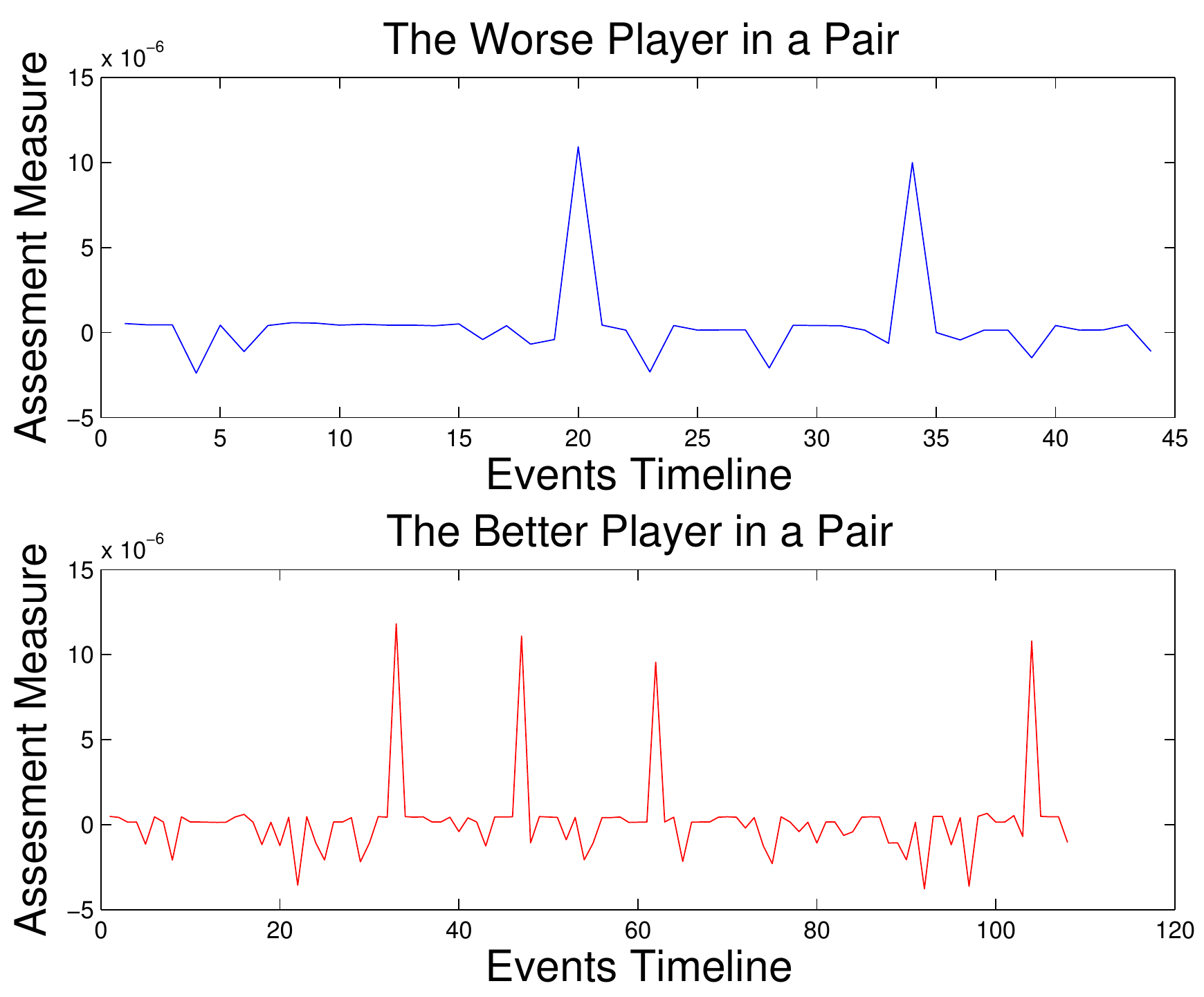}}~
\subfigure[A Pair of Players  \#3]{\label{good_spatial_fig}\includegraphics[height=0.15\textheight]{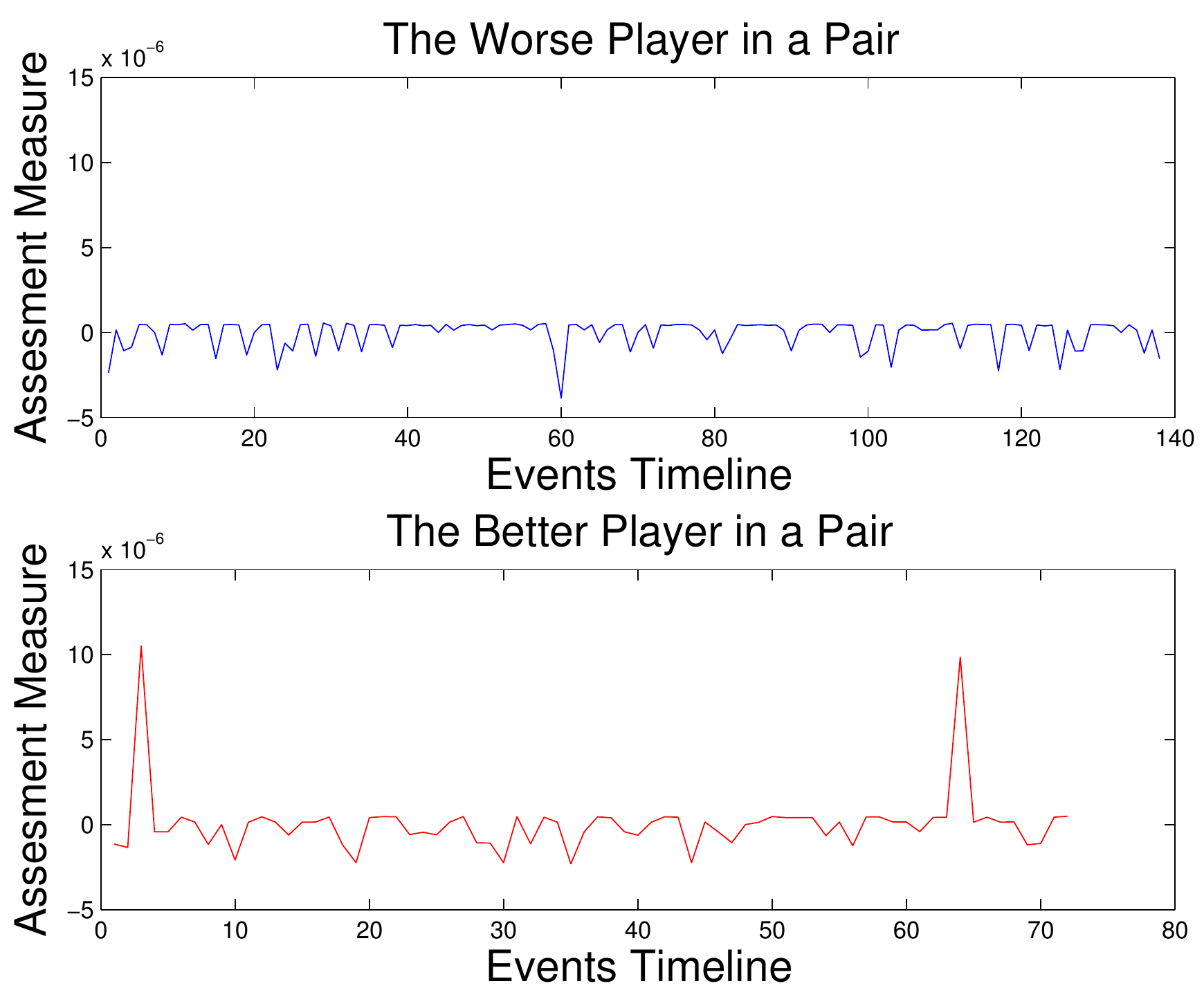}}~
\subfigure[A Pair of Players  \#4]{\label{bad_spatial_fig}\includegraphics[height=0.15\textheight]{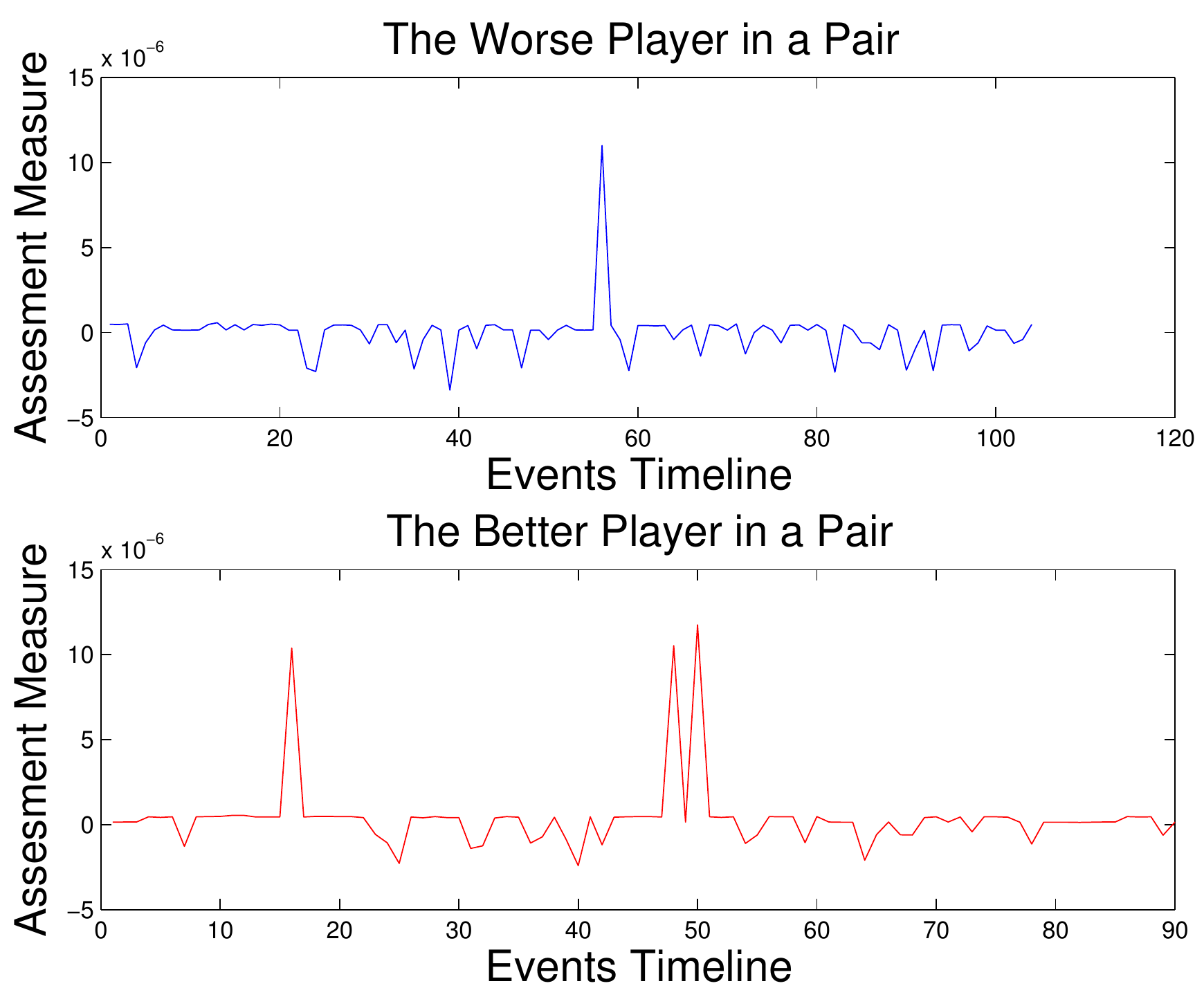}}~
\vspace{-0.4cm}
\caption{We randomly select $4$ pairs of basketball players, and visualize how our assessment model evaluates each player over time. The red plot denotes the better player in a pair, whereas the blue plot depicts the worse player. The $y$-axis in the plot illustrates our predicted performance measure for an event occurring at a specific time in a player's first-person video.\vspace{-0.4cm}}
    \label{dynamic_fig}
\end{figure*}

\section{Experimental Results}

\subsection{Quantitative Results}

\textbf{Atomic Basketball Event Detection.} In Table~\ref{att_table}, we first illustrate our results for atomic basketball event detection task.  The results are evaluated according to the maximum F-score (MF) metric by thresholding the predicted atomic event probabilities at small intervals and then computing a precision and recall curve. First, we compare our model's predictions with several recent first-person activity recognition baselines~\cite{Singh_2016_CVPR,gberta_2017_RSS,ma2016going} and also with the successful video activity recognition baseline C3D~\cite{Tran:2015:LSF:2919332.2919929}. We show that our model outperforms all of these baselines for each atomic event.

Furthermore, to justify our model's design choices, in Table~\ref{att_table} we also include several experiments studying the effect of 1) a multi-path architecture, 2) LSTM layers, and 3) zooming-in scheme. Our experiments indicate that each of these components is crucial for achieving a solid atomic event recognition accuracy, i.e. the system achieves the best performance when all three of these components are included in the model.

  \begin{table}
   \footnotesize
    \begin{center}
    \begin{tabular}{ c  | c  | c |}
    \cline{2-3}
    & \multicolumn{2}{|c |}{Accuracy} \\
     \cline{2-3}
     &  \multicolumn{1}{ c |}{Pred. Events} &  \multicolumn{1}{ c |}{GT Events} \\ \cline{1-3}
    	  \multicolumn{1}{| c |}{LRCN~\cite{lrcn2014} 2-pt made shot detector} & fail & - \\	
    	  \multicolumn{1}{| c |}{ LRCN~\cite{lrcn2014} 3-pt made shot detector} & fail & - \\ \hline
    	  \multicolumn{1}{| c |}{Ours: no GMs} & 0.477 & -\\ 
	   \multicolumn{1}{| c |}{Ours: no $p^{(3)}$} & 0.496 & -\\
	   \multicolumn{1}{| c |}{Ours: no $p^{(2)}$} & 0.515 & -\\
	  \multicolumn{1}{| c |}{ Ours: no $p^{(1)}$} & 0.536 & -\\
	  \multicolumn{1}{| c |}{Ours: single GM-top2} & 0.537 & - \\
	   \multicolumn{1}{| c |}{Ours: all weights $w$ set to $1$} & 0.583 & -\\ 
	   \multicolumn{1}{| c |}{Ours: single GM-top1} & 0.609 &  - \\
	   \multicolumn{1}{| c |}{Ours: no $p^{(4)}$} & 0.649 & -\\  \hline
	   \multicolumn{1}{| c |}{Ours} & \bf 0.765 & \bf 0.793\\ \hline	
    \end{tabular}
    \end{center}\vspace{-.4cm}
    
    \caption{The quantitative results for our basketball assessment task. We evaluate our method on $250$ labeled pairs of players, and predict, which of the two players in a pair is better. We then compute the accuracy as the fraction of correct predictions. We report the results of various baselines in two settings: 1) using our predicted atomic events, and 2) using ground truth atomic events. These results show that 1) our model achieves best results, 2) that each of our proposed components is important, and 3) that our system is pretty robust to atomic event recognition errors. \vspace{-0.4cm}}
    \label{skill_table}
   \end{table}

   



\textbf{Basketball Assessment Results.} In Table~\ref{skill_table}, we present our results  for assessing $24$ basketball players from our testing dataset. To test our method's accuracy we evaluate our method on $250$ labeled pairs of players, where a label provided by a basketball expert indicates, which of the two players is better. For each player, our method produces an assessment measure indicating, which player is better (the higher the better). To obtain the accuracy, we compute the fraction of correct predictions across all $250$ pairs.


We note that to the best of our knowledge, we are the first to formally investigate a basketball performance assessment task from a first-person video. Thus, there are no well established prior baselines for this task. As a result, we include the following list of baselines for a comparison.

First, we include two basketball activity baselines: the detectors of 1) 2-point and 2) 3-point shots made by the camera wearer. We label all instances in our dataset where these activities occur and discover $\approx 100$ of such instances. Note that such a small number of instances is not a flaw of our dataset, but instead an inherent characteristic of our task. Such basketball activities belong to a long-tail data distribution, i.e. they occur pretty rarely, and thus, it is difficult to train supervised classifiers for such activity recognition.  We then train an LRCN~\cite{lrcn2014} model as 1) a 2 point made shot detector, and 2) a 3 point made shot detector. We report that due to a small amount of training data, in all cases the network severely overfit the training data and did not learn any meaningful pattern.

Furthermore, to justify each of our proposed components in the model, in Table~\ref{skill_table} we also include several ablation baselines. First, we study how 1) Gaussian Mixtures (GM) and 2) the process of learning the weights affect the performance assessment accuracy. We do it 1) with our predicted and 2) with the ground truth atomic events. We show that in both cases, each of our proposed components is beneficial. In addition, we also observe that our system is robust to atomic event recognition errors: the accuracy when using the ground truth atomic events is only $2.8\%$ better compared to our original model.


We also present the performance assessment results when we remove one of the four atomic events from our system. We show that our method performs the best when all four atomic events are used, suggesting that each atomic event is useful. Finally, as two extra baselines we manually select two Gaussian mixtures with the largest weight magnitudes and use each of their predictions independently (denoted as single GM-top1,2 in Table~\ref{skill_table}). We show that our full model outperforms all the other baselines, thus, indicating that each of our proposed component in our model is crucial for an accurate player performance assessment. 



\captionsetup{labelformat=empty}
\captionsetup[figure]{skip=5pt}

\begin{figure*}
\centering

\myfiguresixcol{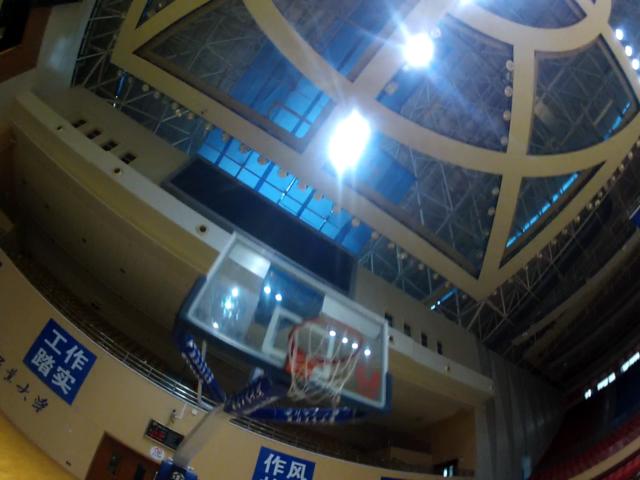}
\myfiguresixcol{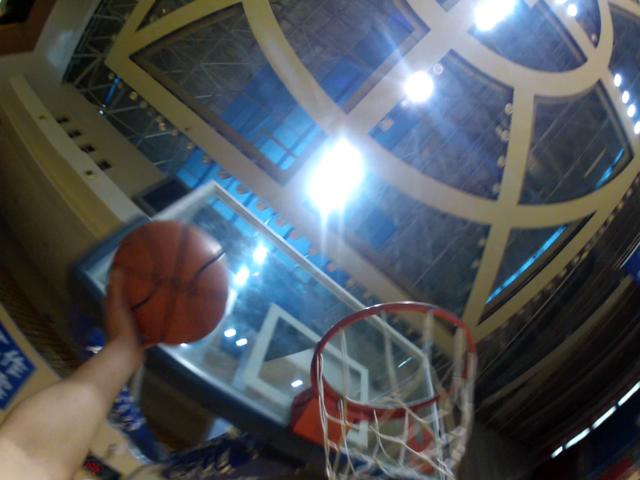}
\myfiguresixcol{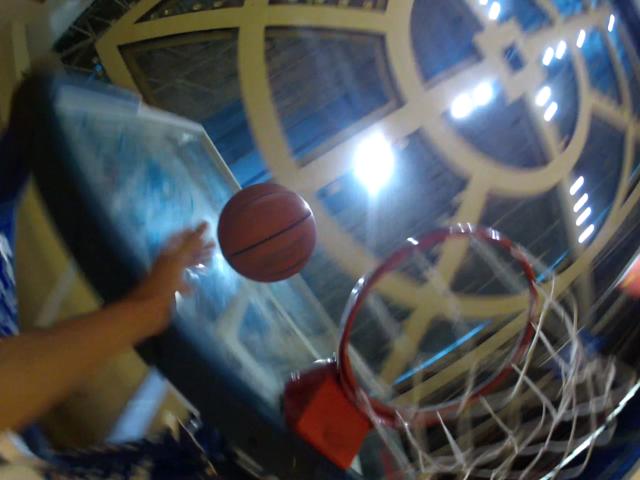}
\myfiguresixcol{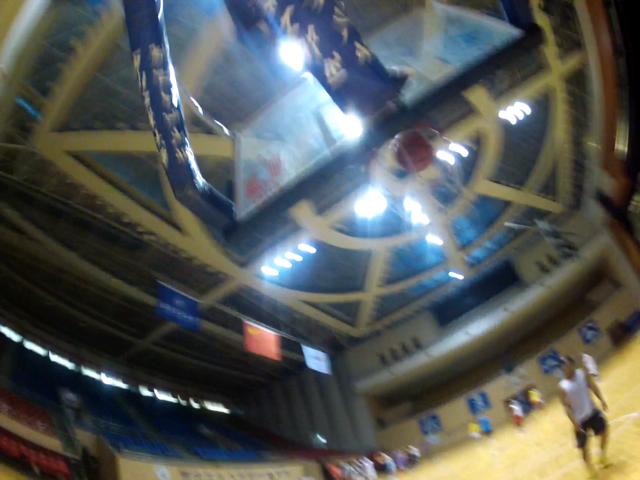}
\myfiguresixcol{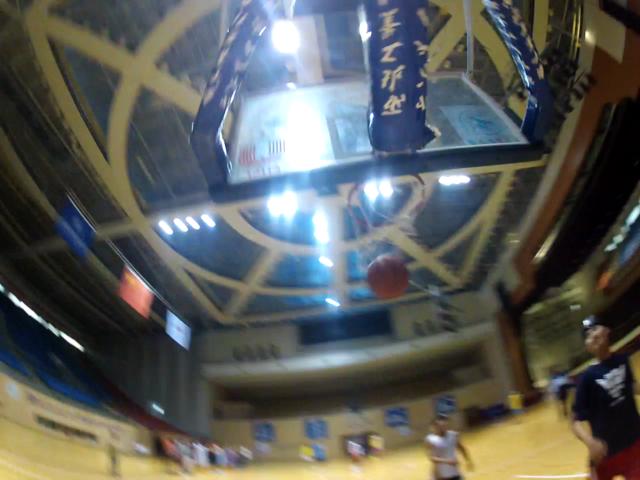}
\myfiguresixcol{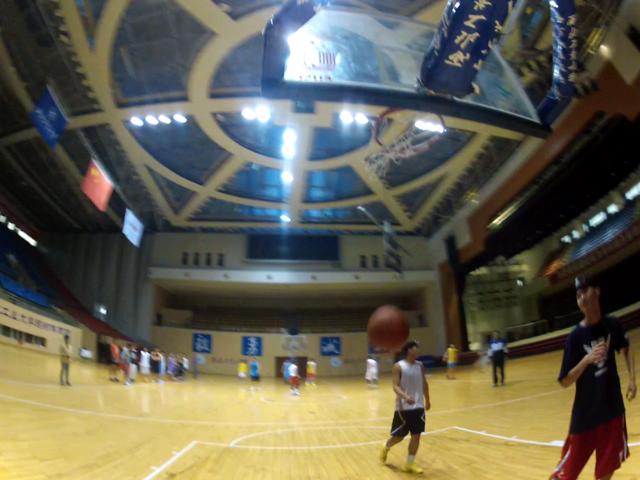}

\myfiguresixcol{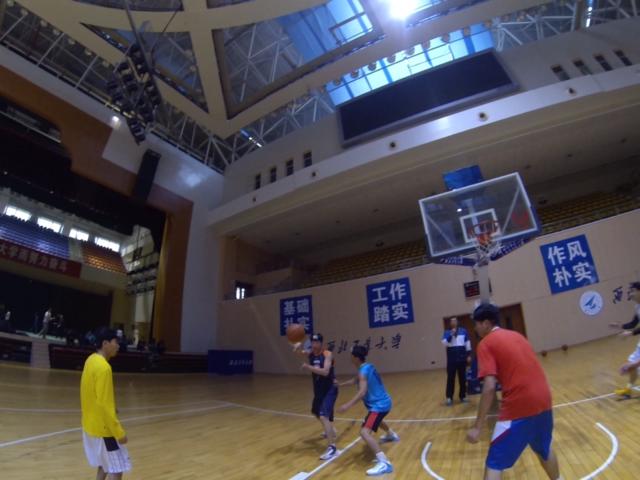}
\myfiguresixcol{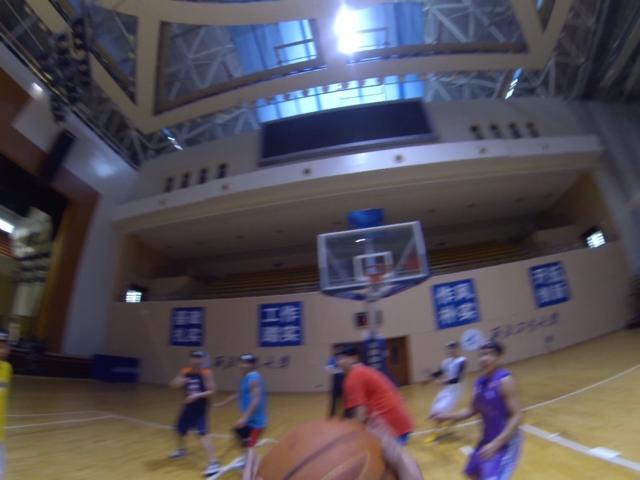}
\myfiguresixcol{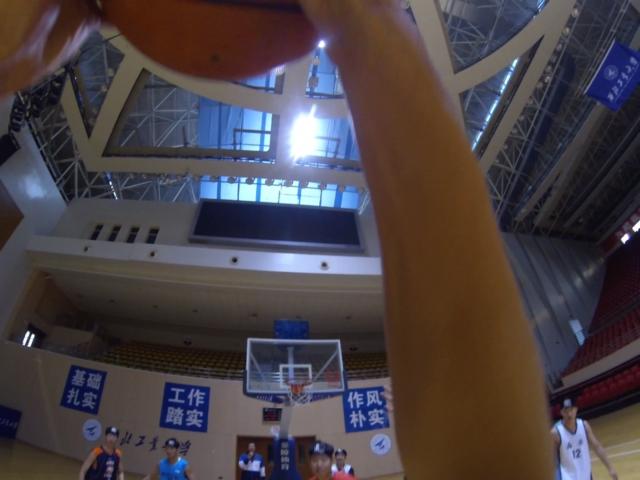}
\myfiguresixcol{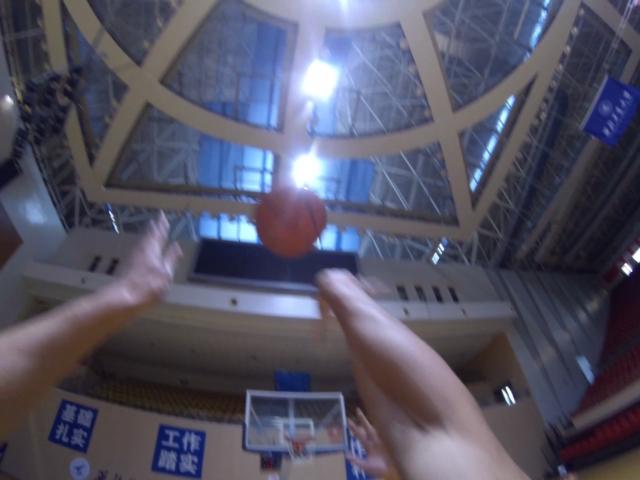}
\myfiguresixcol{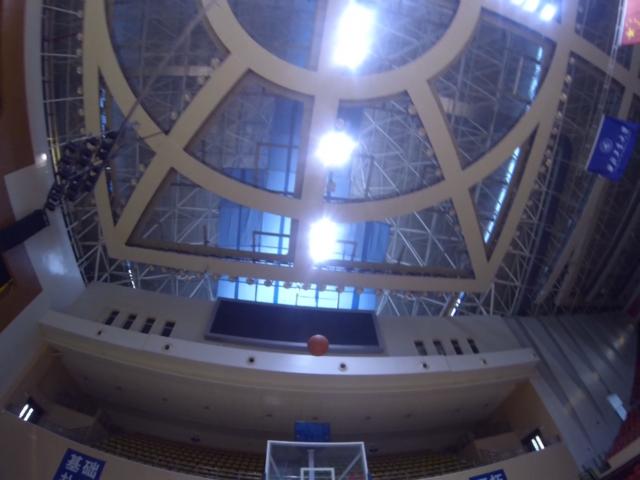}
\myfiguresixcol{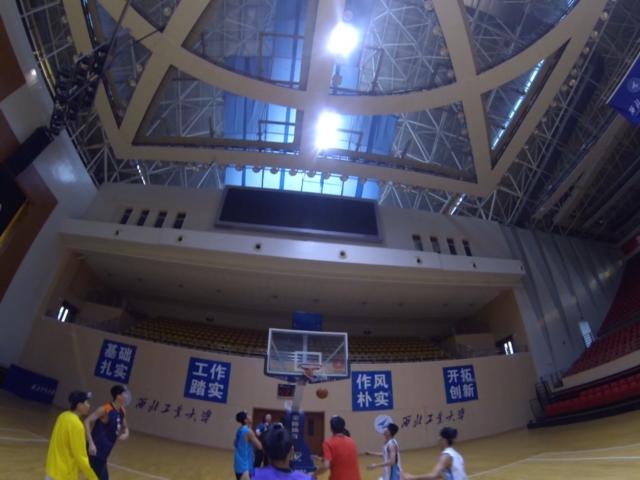}

\myfiguresixcol{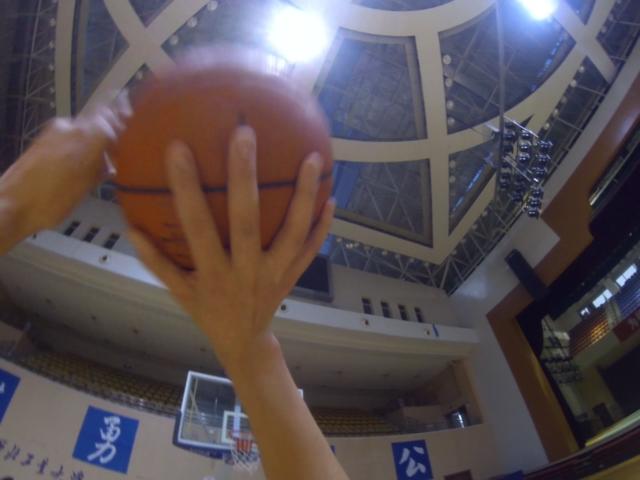}
\myfiguresixcol{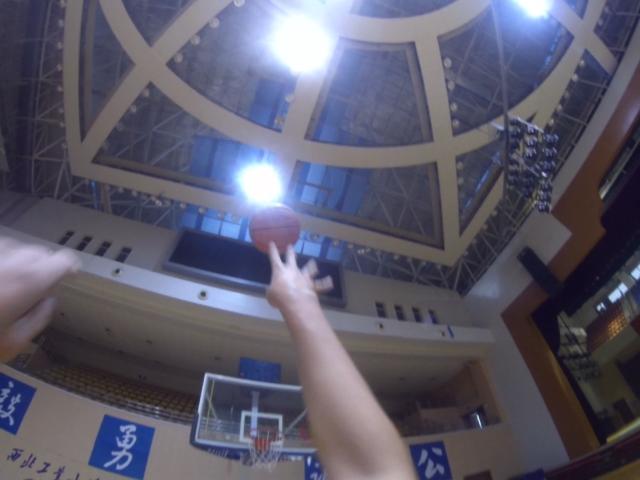}
\myfiguresixcol{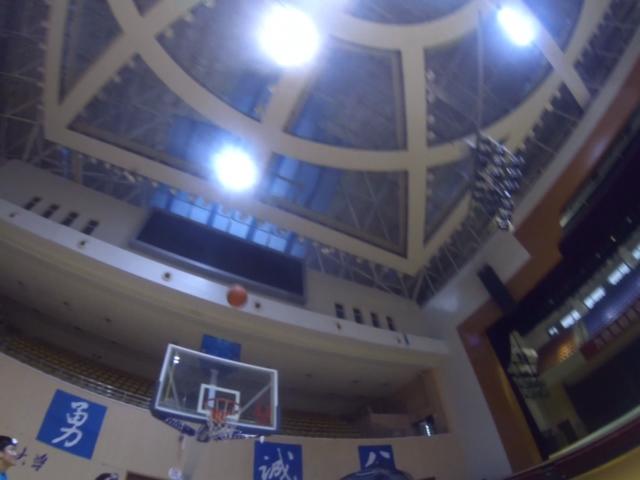}
\myfiguresixcol{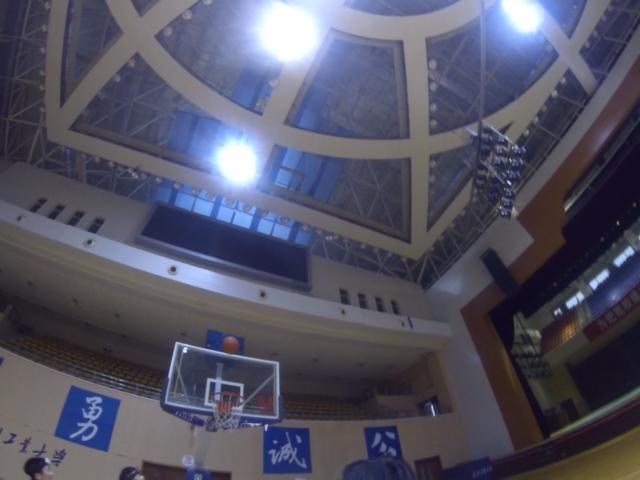}
\myfiguresixcol{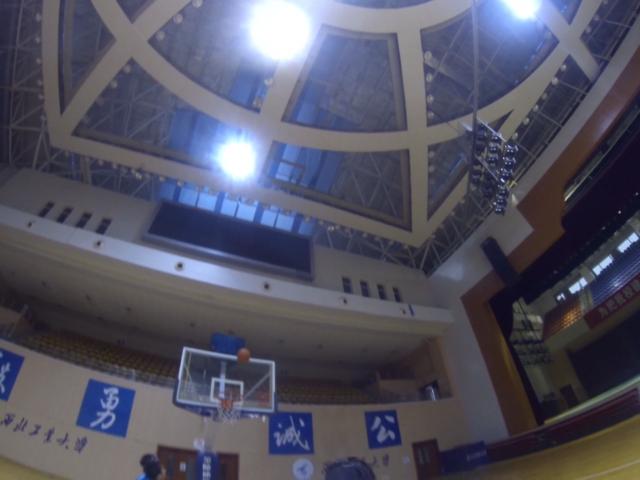}
\myfiguresixcol{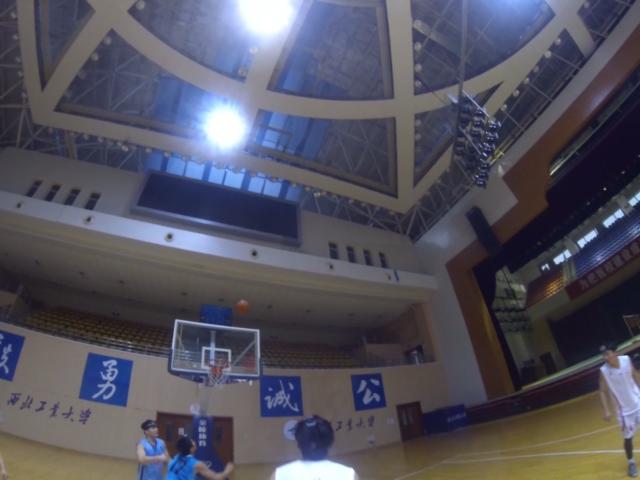}

\captionsetup{labelformat=default}
\setcounter{figure}{4}
\caption{A visualization of basketball activities that we discovered by manually inspecting Gaussian mixtures associated with the largest basketball assessment model weights $w$. Each row in the figure depicts a separate event, and the columns illustrate the time lapse of the event (from left to right), We discover that the two most positive Gaussian mixtures correspond to the events of a player making a 2 point and a 3 point shot respectively (the first two rows), while the mixture with the most negative weight captures an event when a player misses a 2 point shot (last row). \vspace{-0.5cm}}
    \label{gmm_fig}
\end{figure*}

\captionsetup{labelformat=default}
\captionsetup[figure]{skip=10pt}


\subsection{Qualitative Results}

In addition, in Figure~\ref{dynamic_fig}, we also include a more dynamic visualization of how our assessment model works over time. To do this, we randomly select $4$ pairs of basketball players, and visualize how our model evaluates each player over time. The red plot in each pair denotes the better player, whereas the blue plot depicts the worse player. The $y$-axis in the plot illustrates our predicted performance measure for an event occurring at a specific time in a player's first-person video.
 
Furthermore, in Figure~\ref{best_worst_fig} we also include examples of short sequences, illustrating 1) a player's actions that contributed most positively to his/her performance assessment and also 2) actions that contributed most negatively. We select these action sequences by picking the first-person video sequences with a largest positive and negative values of the terms inside the summation of Equation~\ref{Eq:assessment} (which also correspond to positive and negative peaks from Figure~\ref{dynamic_fig}). Such terms depict each video segment's contribution to the overall basketball skill assessment measure.

We would like to note that it is quite difficult to include such results in an image format, because 1) images are static and thus, they cannot capture the full content of the videos; 2) images in the paper, appear at a very low-resolution compared to the original $480 \times 640$ videos, which makes it more difficult to understand what kind of events are depicted in these images. To address some of these issues, in our supplementary material, we include even more of such qualitative examples in a video format.

\textbf{Understanding the Feature Representation.} Earlier, we claimed that Gaussian mixtures produce a highly non-linear feature representation. We now want to get a better insight into what it represents. To do so we analyze the learned weights $w$, and then manually inspect the Gaussian mixtures associated with the largest magnitude weights in $w$. Upon doing so we discover that the two mixtures with the most positive weights learn to capture basketball activities when camera wearer makes a 2 point shot, and a 3 point shot respectively. Conversely, the mixtures with the two most negative weights represent the activities of the camera missing a 2 point shot, and the camera wearer's defender making a shot respectively. In Figure~\ref{gmm_fig}, we include several sequences corresponding to such discovered activities.


\section{Conclusions}

In this work, we introduced a basketball assessment model that evaluates a player's performance from his/her first-person basketball video. We showed that we can learn powerful visual spatiotemporal assessment features from first-person videos, and then use them to learn our skill assessment model from the pairs of weakly labeled first-person basketball videos. We demonstrated that despite not knowing the labeler's assessment criterion, our model learns to evaluate players with a solid accuracy. In addition, we can also use our model to discover the camera wearer's activities that contribute positively or negatively to his/her performance assessment.

We also note that performance assessment is an important problem in many different areas not just basketball. These include musical instrument playing, job related activities, and even our daily moments such as cooking a meal. In our future work, we plan to investigate these new areas, and try to generalize our model to such activities too.

\captionsetup{labelformat=empty}
\captionsetup[figure]{skip=5pt}

\begin{figure*}
\centering

\myfiguresixcol{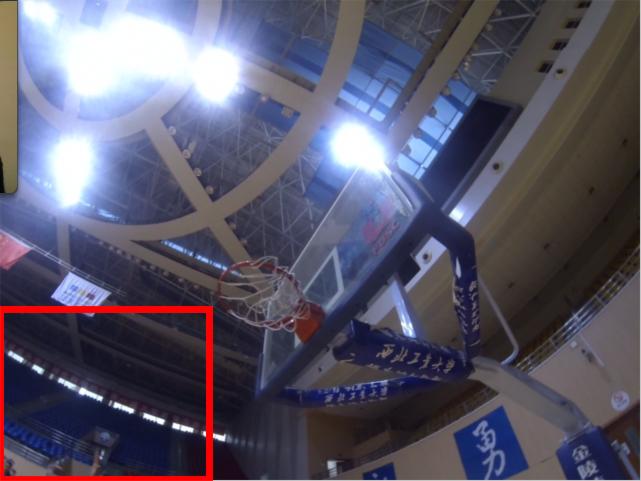}
\myfiguresixcol{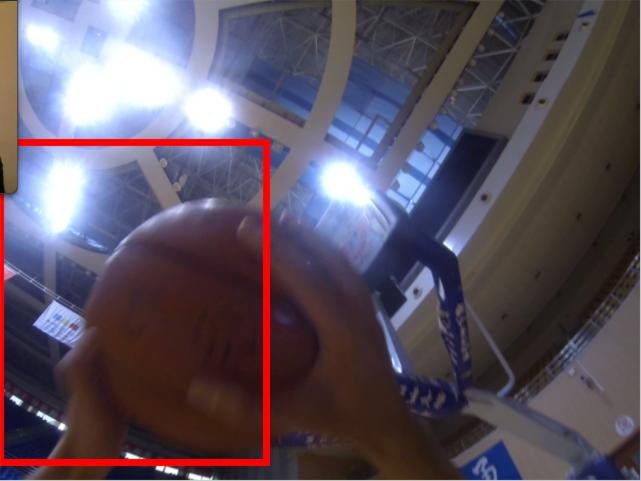}
\myfiguresixcol{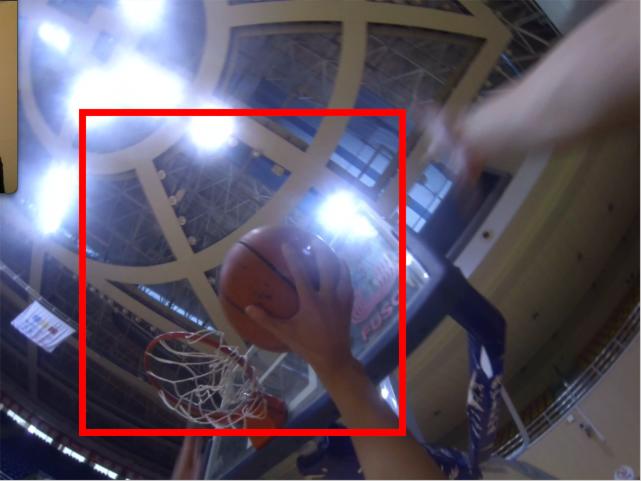}
\myfiguresixcol{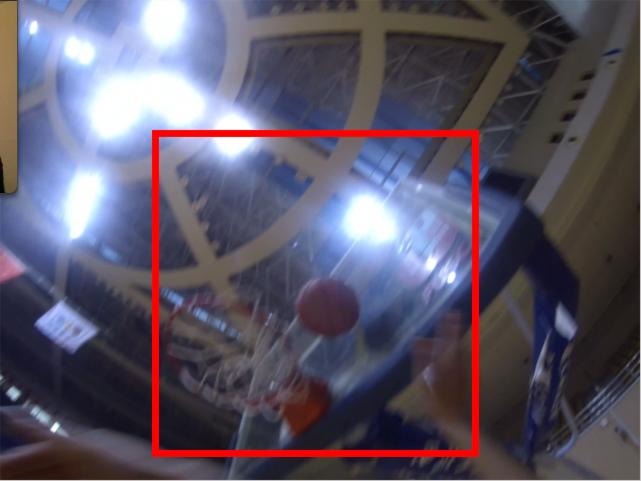}
\myfiguresixcol{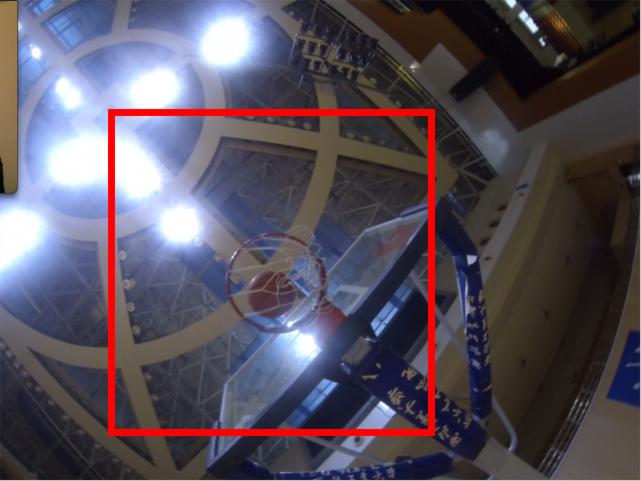}
\myfiguresixcol{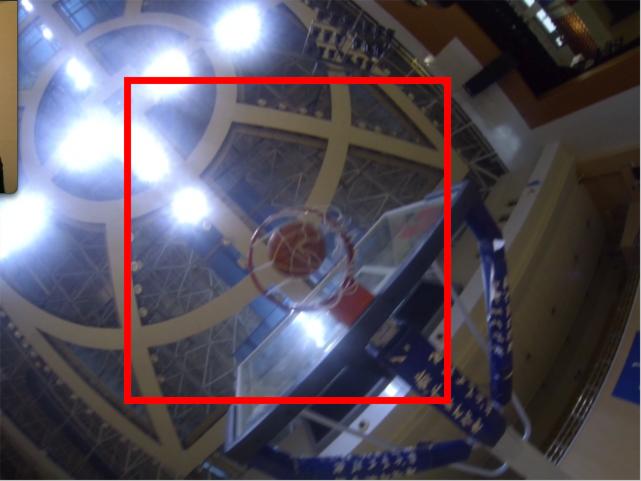}

\myfiguresixcol{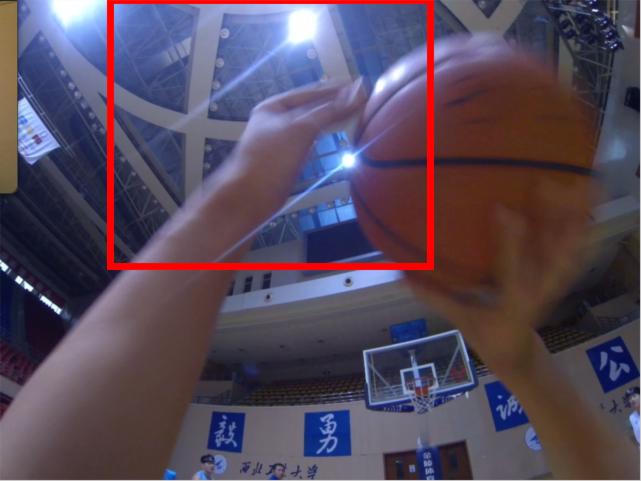}
\myfiguresixcol{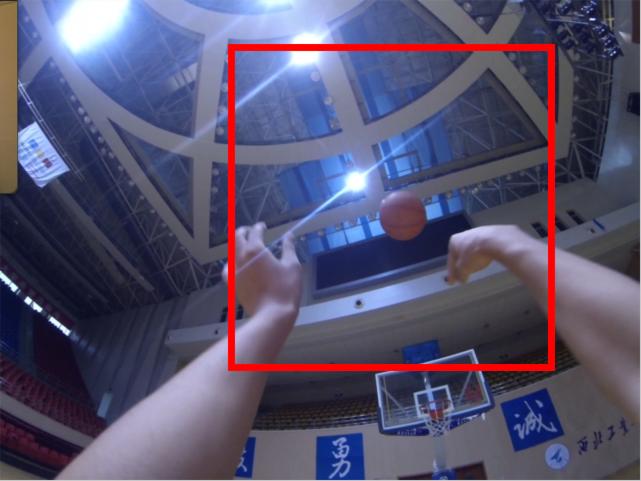}
\myfiguresixcol{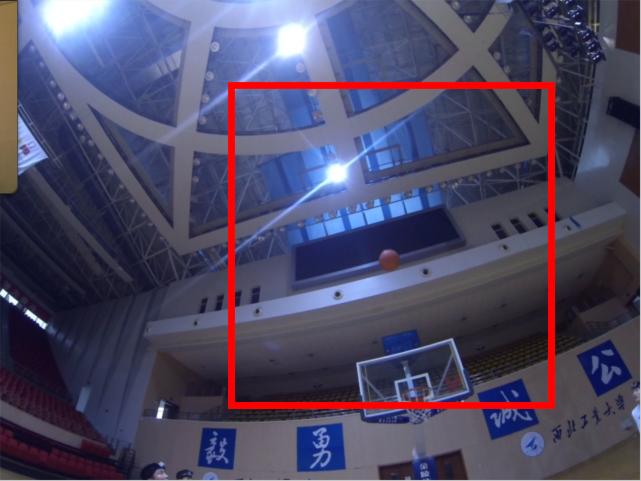}
\myfiguresixcol{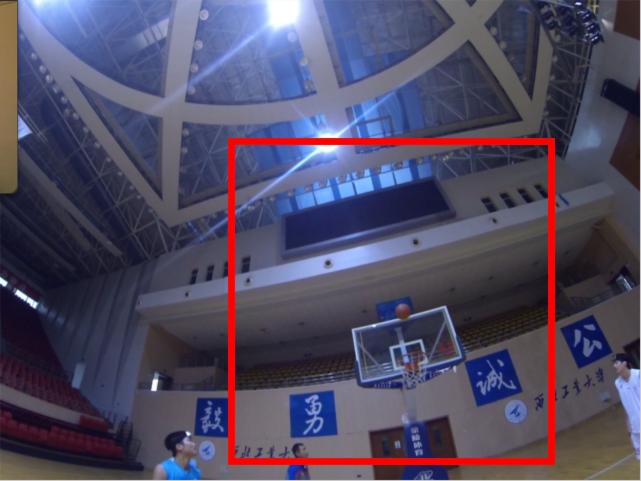}
\myfiguresixcol{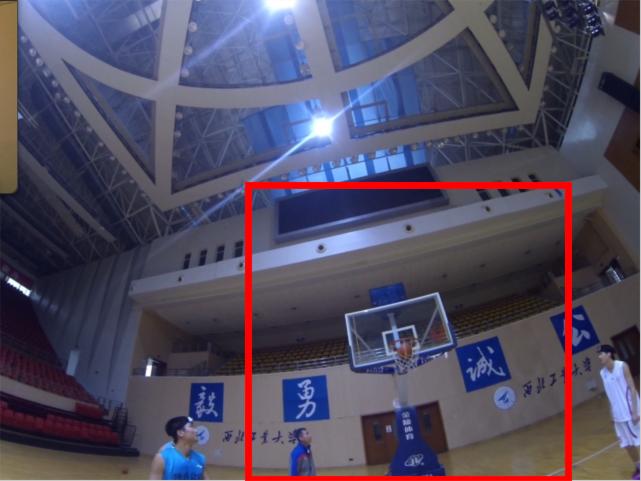}
\myfiguresixcol{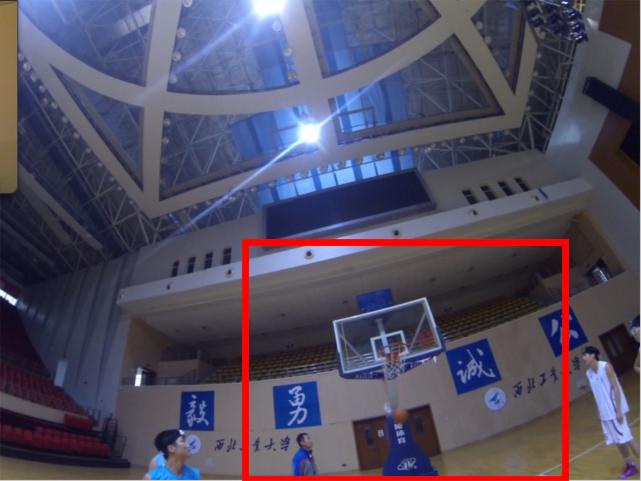}

\myfiguresixcol{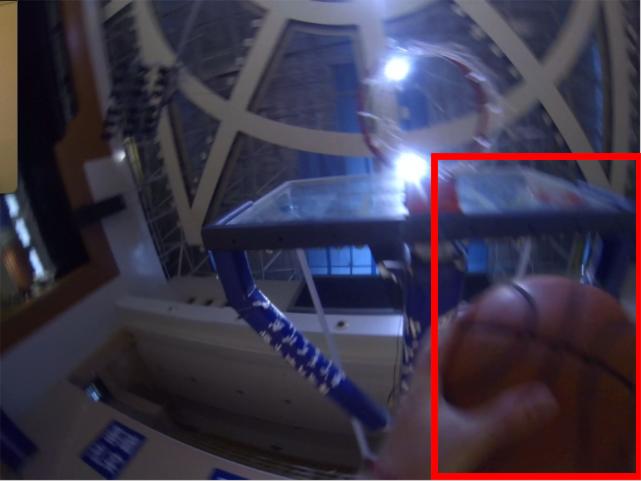}
\myfiguresixcol{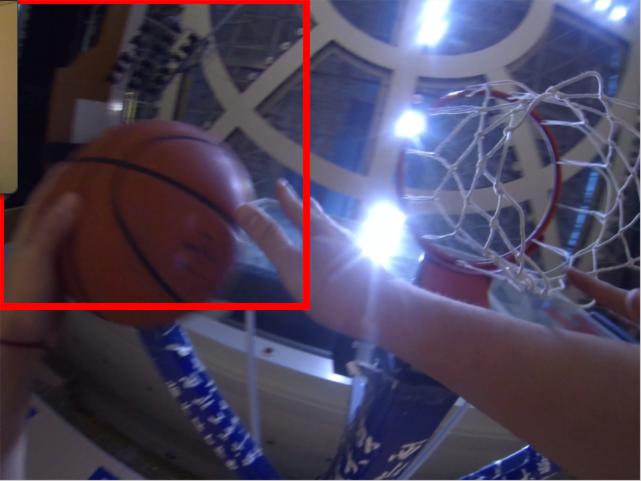}
\myfiguresixcol{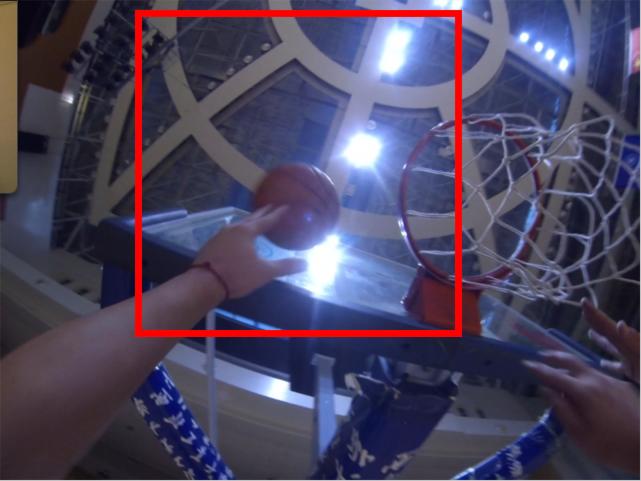}
\myfiguresixcol{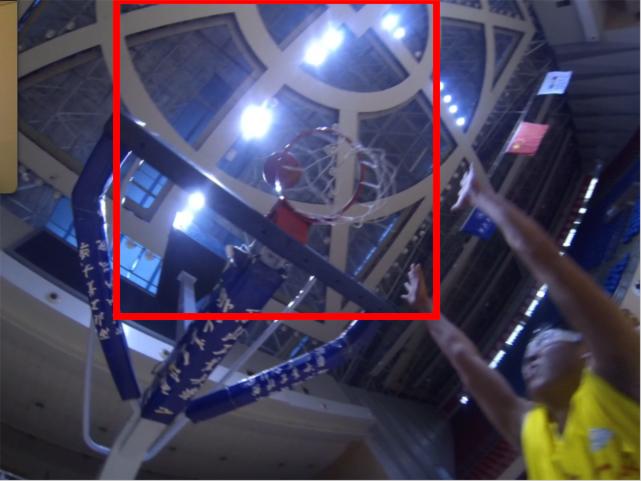}
\myfiguresixcol{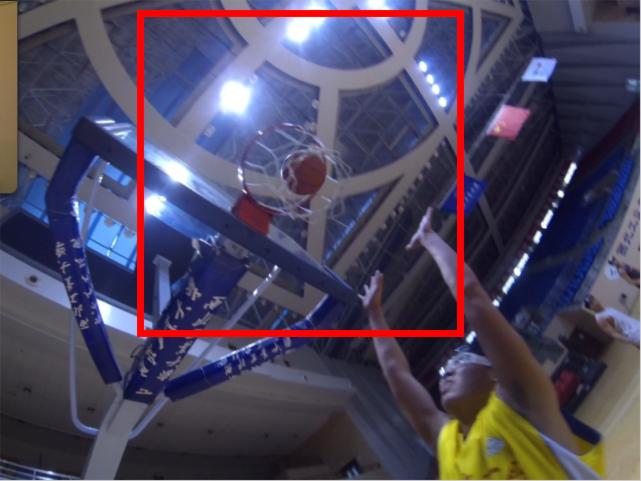}
\myfiguresixcol{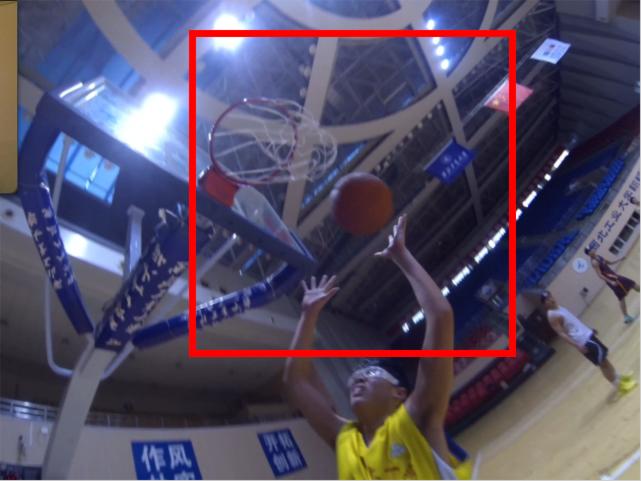}

\myfiguresixcol{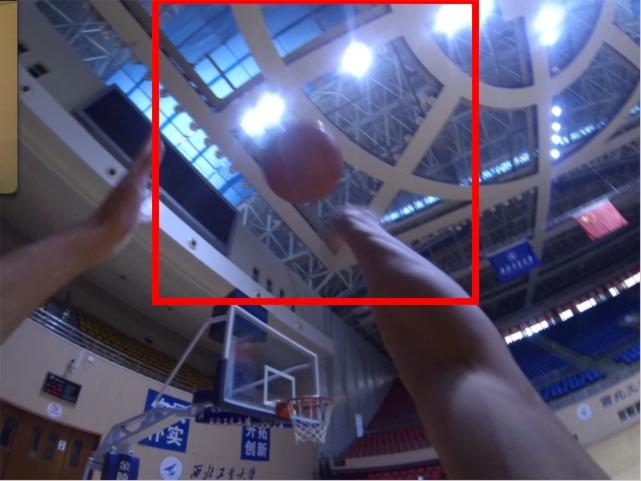}
\myfiguresixcol{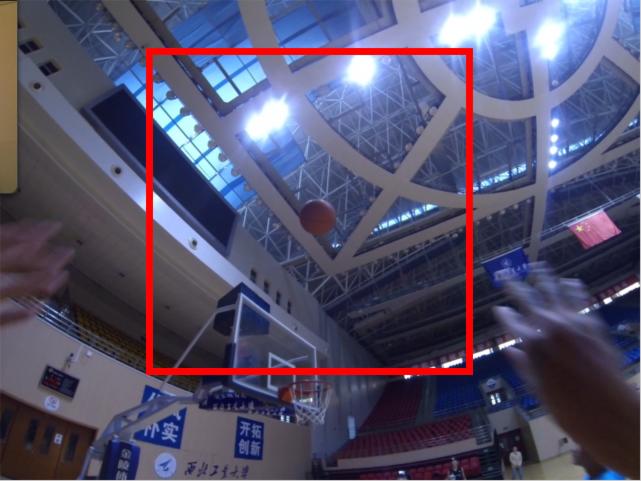}
\myfiguresixcol{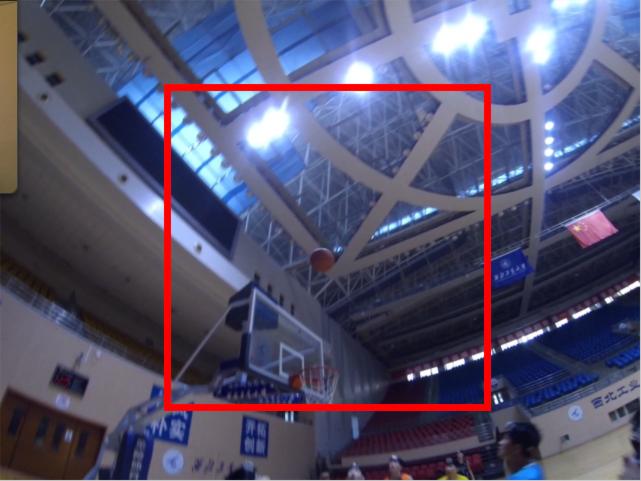}
\myfiguresixcol{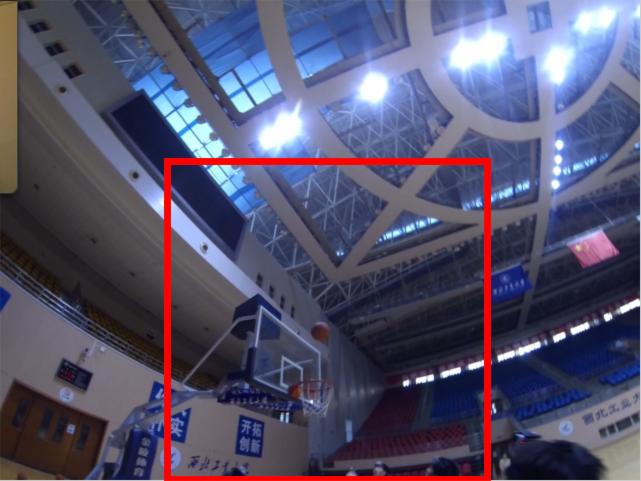}
\myfiguresixcol{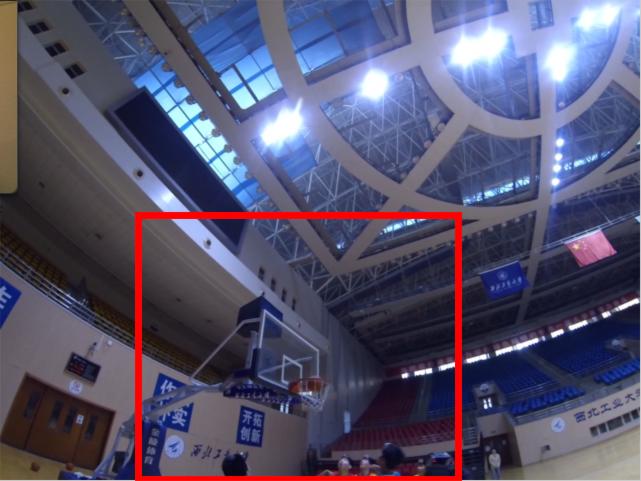}
\myfiguresixcol{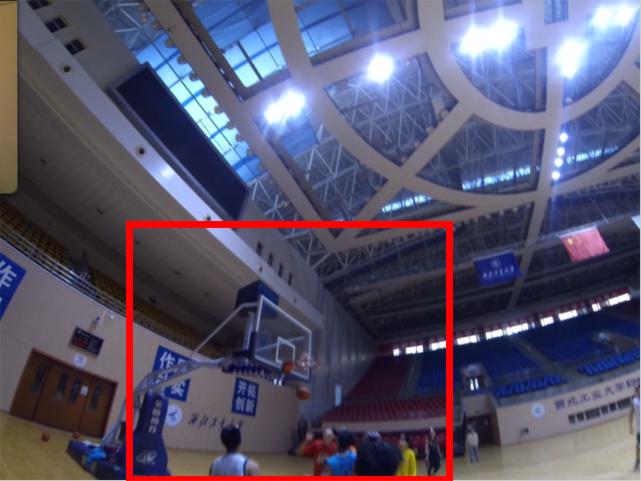}

\myfiguresixcol{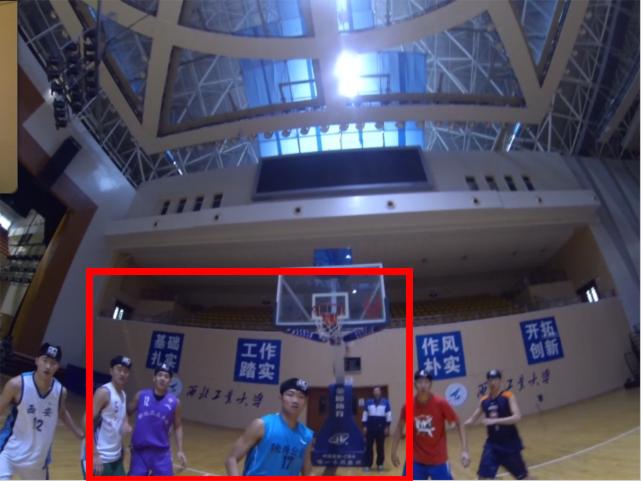}
\myfiguresixcol{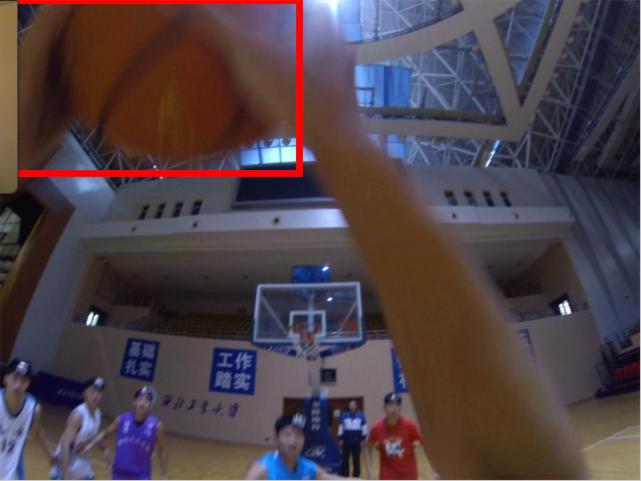}
\myfiguresixcol{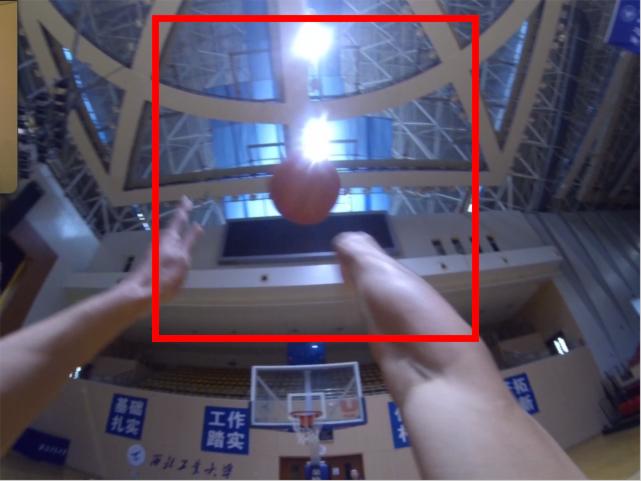}
\myfiguresixcol{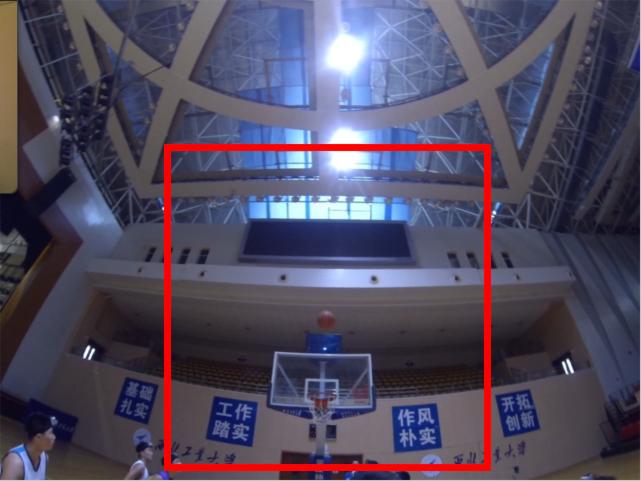}
\myfiguresixcol{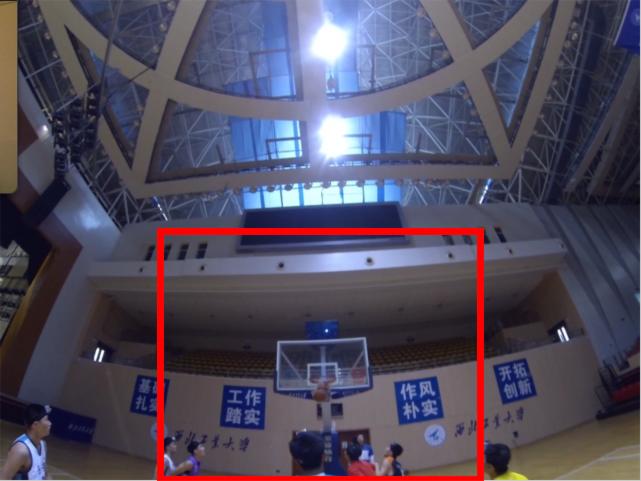}
\myfiguresixcol{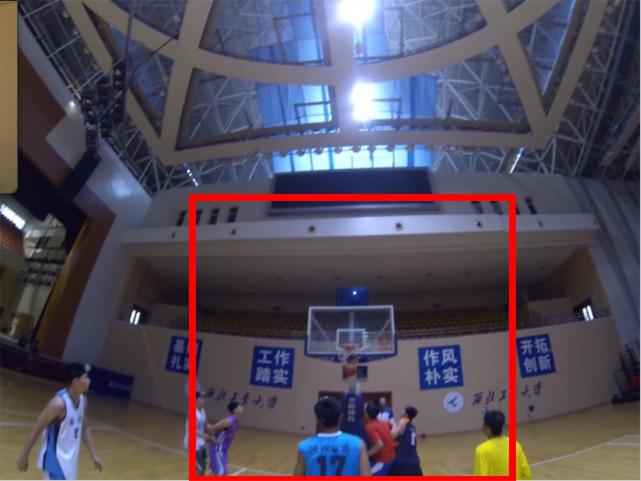}

\vspace{-0.1cm}
\small{(a) The detected events that contributed most \textbf{positively} to a player's performance assessment score according to our model}
\normalsize{}
\vspace{0.2cm}


\myfiguresixcol{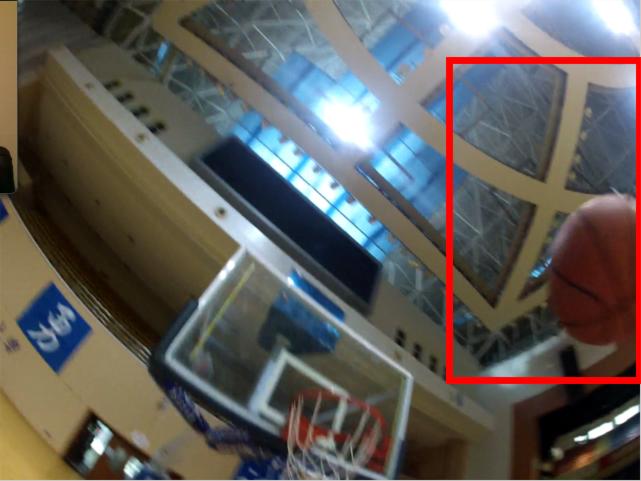}
\myfiguresixcol{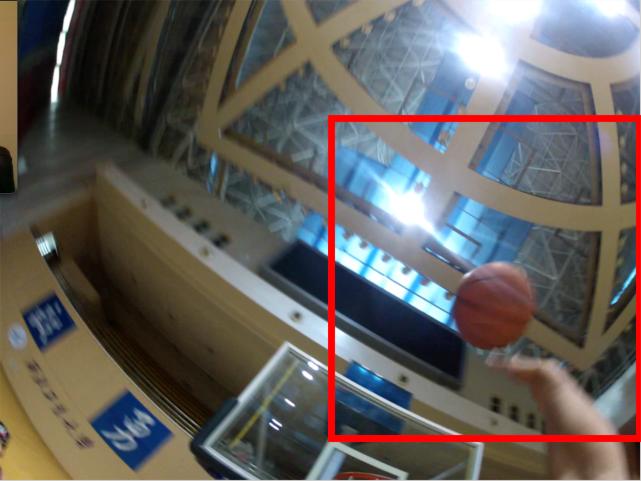}
\myfiguresixcol{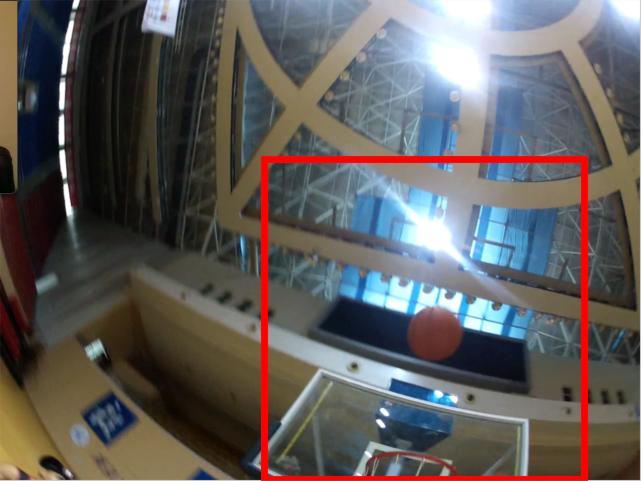}
\myfiguresixcol{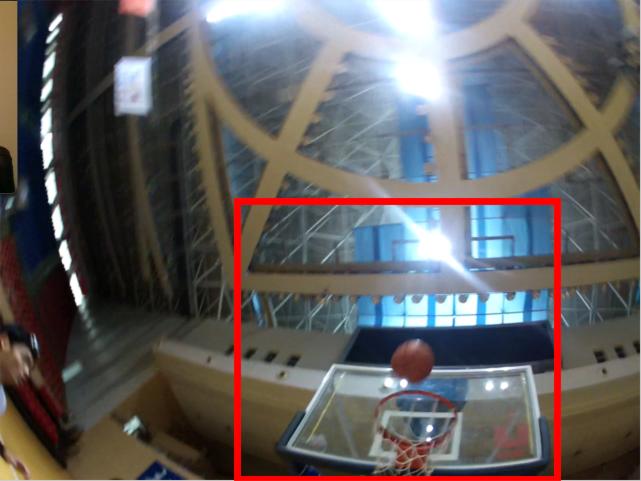}
\myfiguresixcol{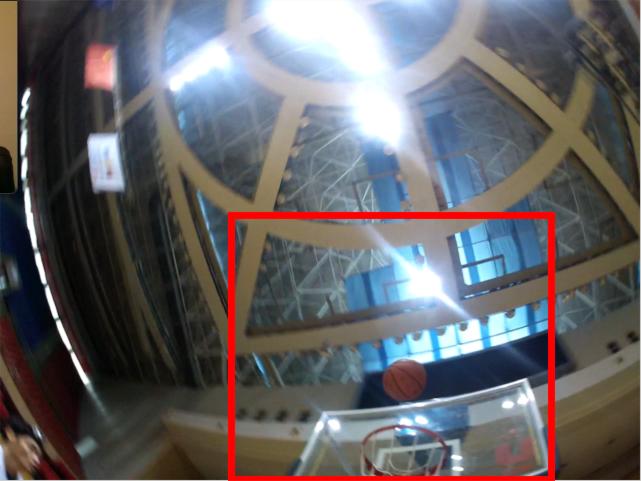}
\myfiguresixcol{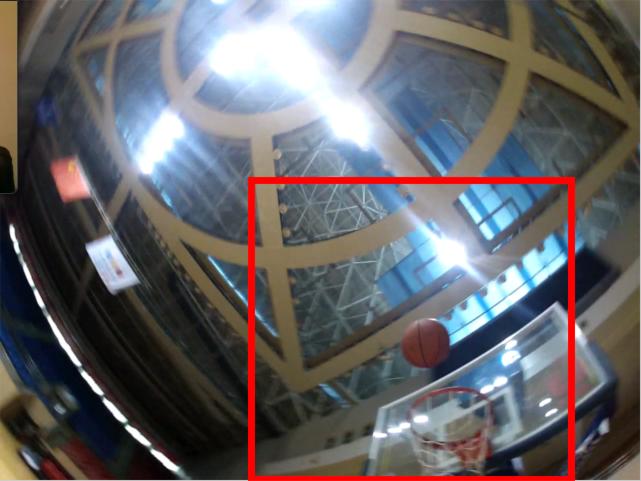}

\myfiguresixcol{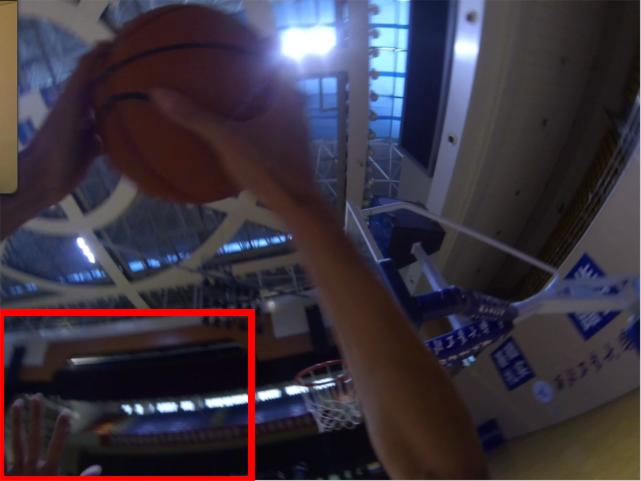}
\myfiguresixcol{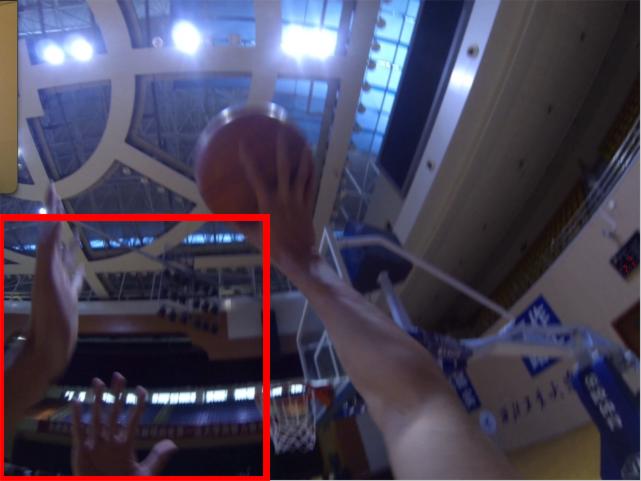}
\myfiguresixcol{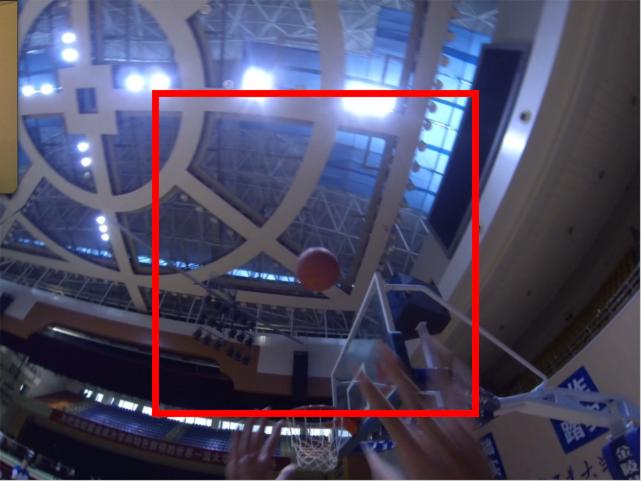}
\myfiguresixcol{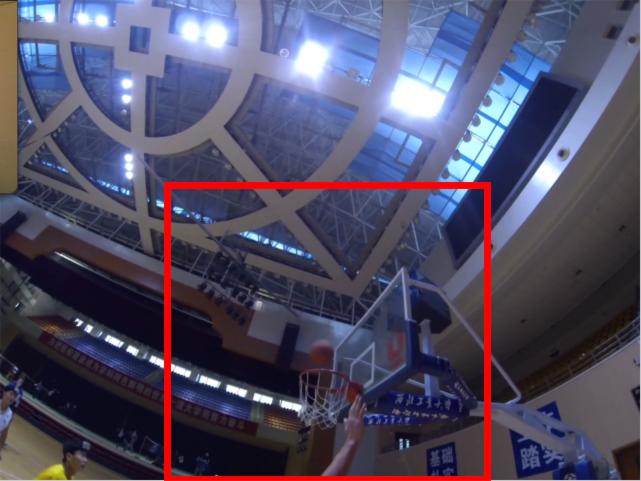}
\myfiguresixcol{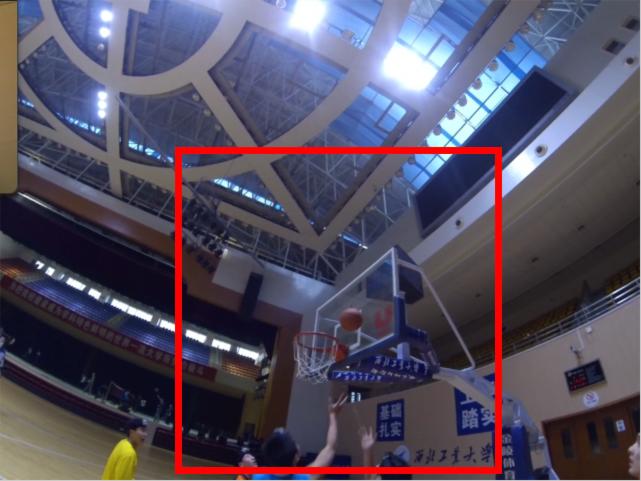}
\myfiguresixcol{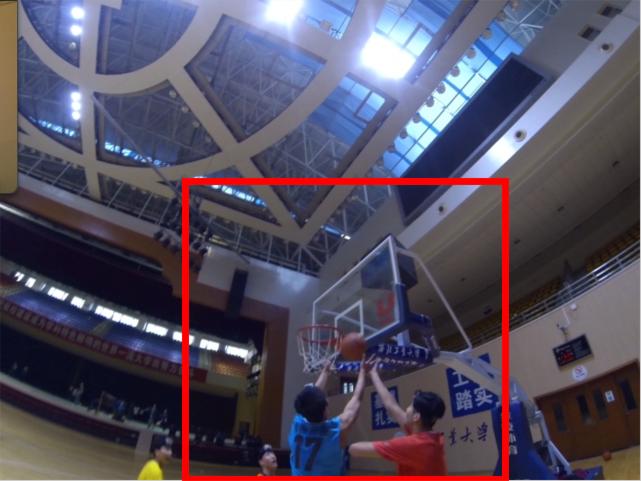}

\myfiguresixcol{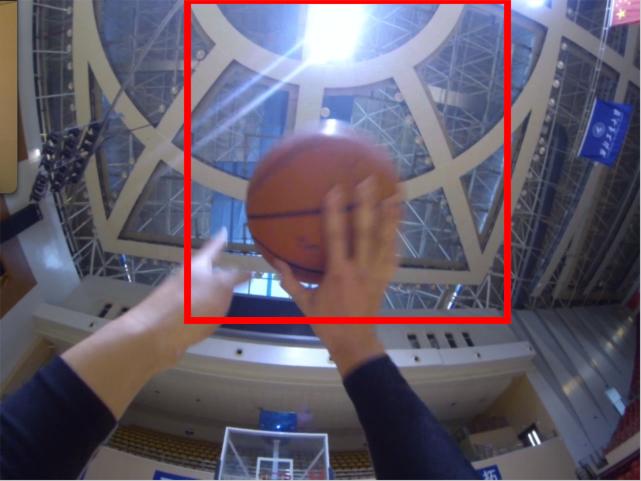}
\myfiguresixcol{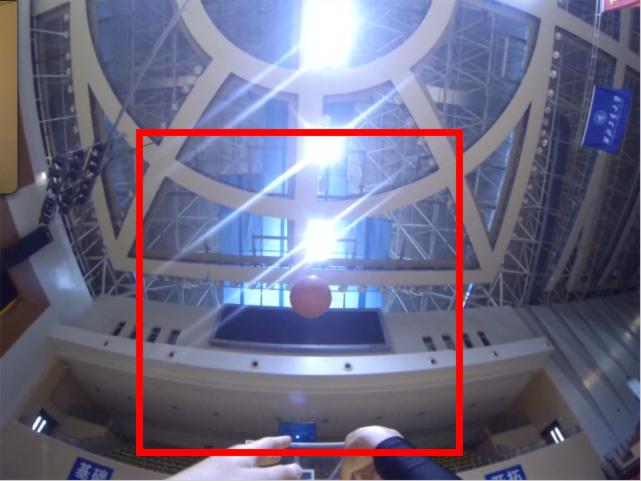}
\myfiguresixcol{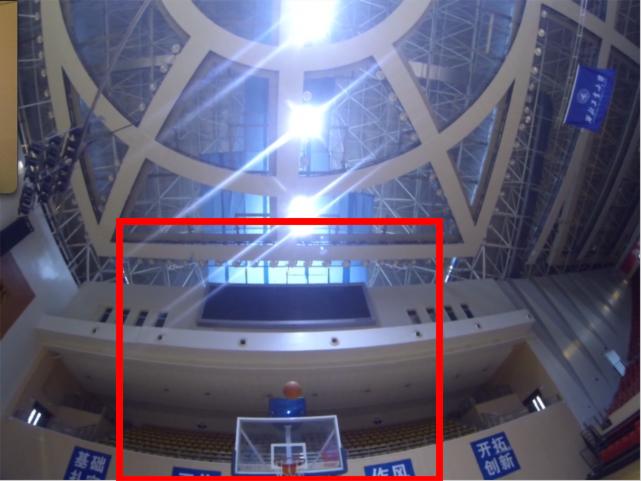}
\myfiguresixcol{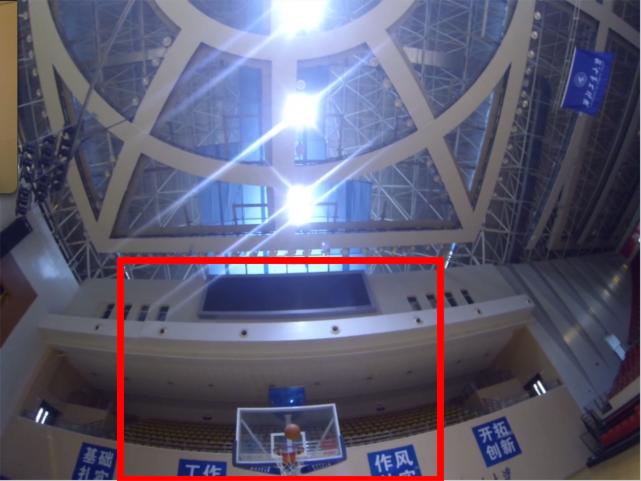}
\myfiguresixcol{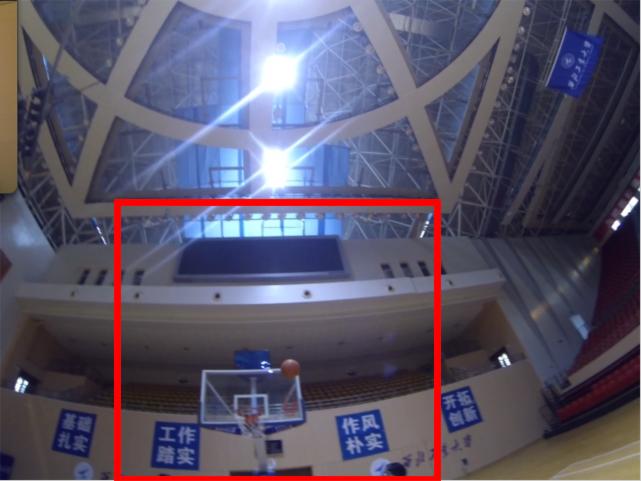}
\myfiguresixcol{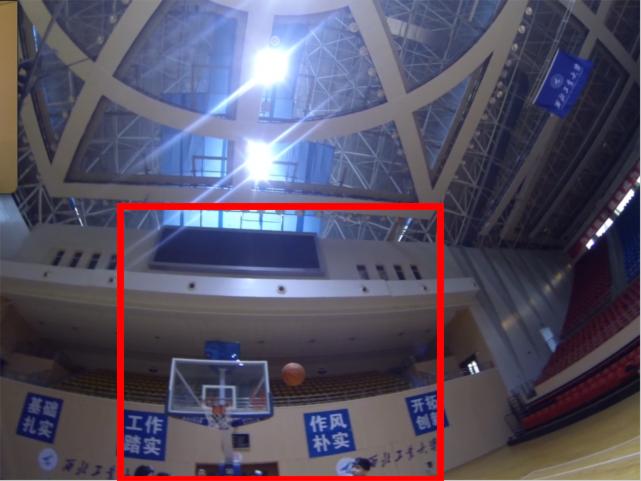}

\myfiguresixcol{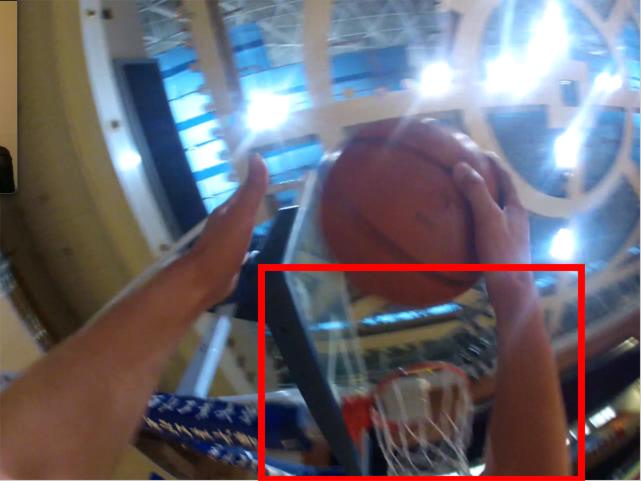}
\myfiguresixcol{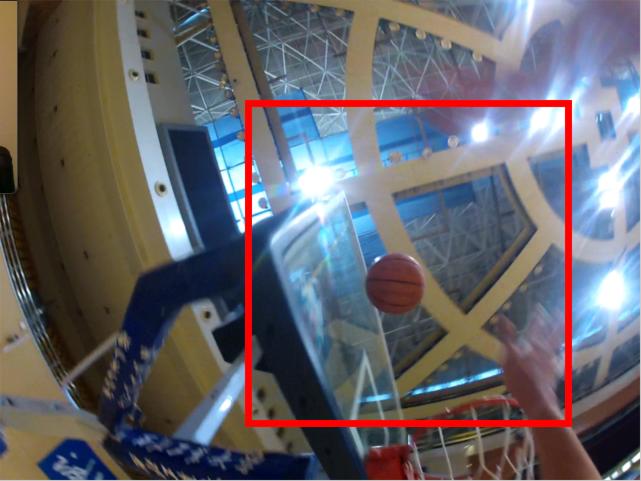}
\myfiguresixcol{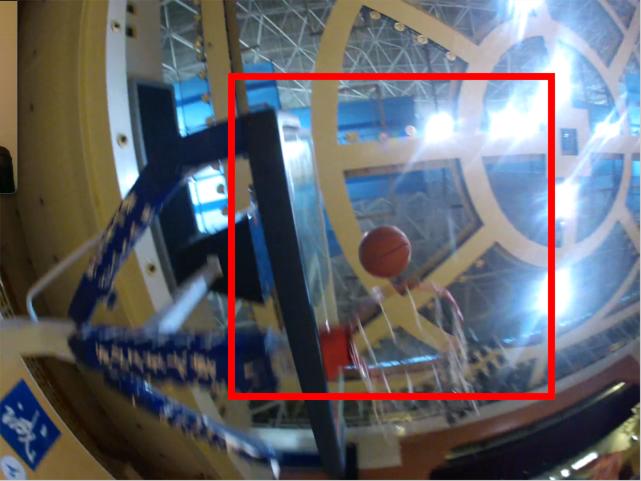}
\myfiguresixcol{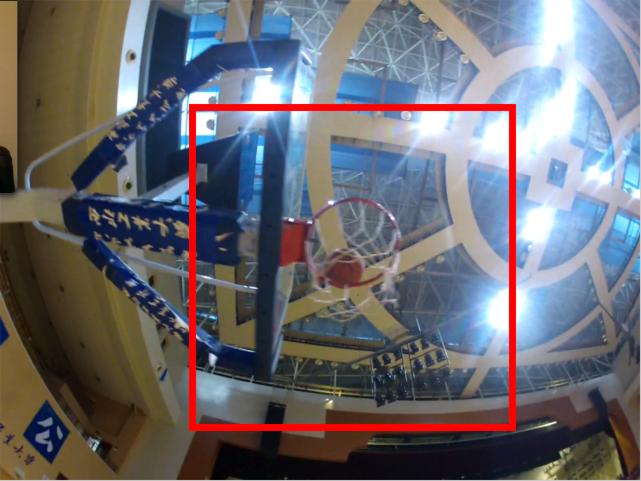}
\myfiguresixcol{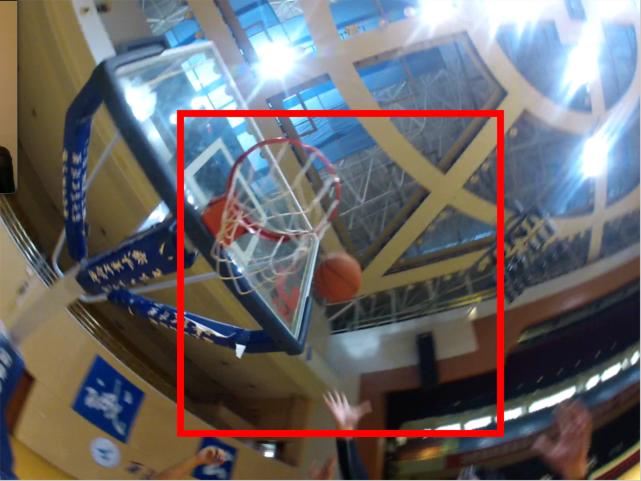}
\myfiguresixcol{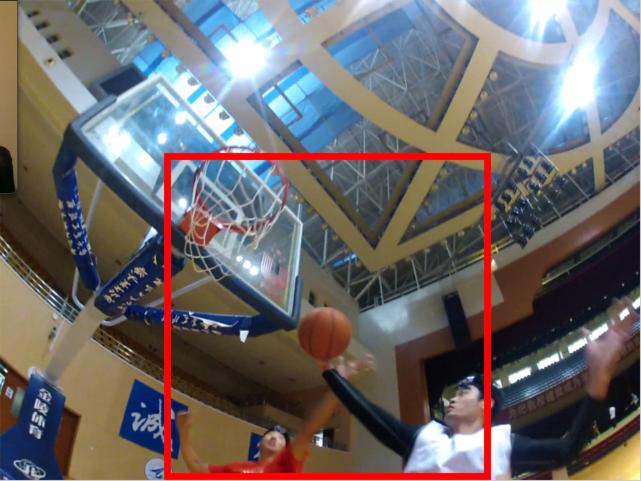}

\vspace{-0.1cm}
\small{(b) The detected events that contributed most \textbf{negatively} to a player's performance assessment score according to our model}
\normalsize{}\vspace{0.1cm}

\captionsetup{labelformat=default}
\setcounter{figure}{5}
\vspace{-0.1cm}
\caption{A figure illustrating the events that contribute most positively (top figure) and most negatively (bottom figure) to a player's performance measure according to our model. The red box illustrates the location where our method zooms-in. Each row in the figure depicts a separate event, and the columns illustrate the time lapse of the event (from left to right). We note that among the detected positive events our method recognizes events such as assists, made layups, and made three pointers, whereas among the detected negative events, our method identifies events such as missed layups, and missed jumpshots. We present more of such video examples in the supplementary material.\vspace{-0.5cm}}
    \label{best_worst_fig}
\end{figure*}

\captionsetup{labelformat=default}
\captionsetup[figure]{skip=10pt}

\bibliographystyle{plain}
\footnotesize{
\bibliography{gb_bibliography_v2,bib_hs_v2}}

\end{document}